\documentclass[10pt,journal,compsoc]{IEEEtran}
\ifCLASSOPTIONcompsoc
  \usepackage[nocompress]{cite}
\else
  \usepackage{cite}
\fi
\ifCLASSINFOpdf
\else
\fi
\usepackage{amsmath,amsfonts,amssymb,amsthm}
\usepackage{algorithm,algorithmic}

  \usepackage{paralist}
  \usepackage{graphics} 
  \usepackage{epsfig} 
\usepackage{graphicx}
\graphicspath{ {images/} }
\usepackage{caption}
\usepackage{subcaption}
\usepackage{float}
 \usepackage{epstopdf}
 \usepackage[colorlinks=true,bookmarks=false]{hyperref}
 \usepackage{arydshln}
 \usepackage{multirow}
 \hypersetup{urlcolor=blue, citecolor=red}
\usepackage{hyperref}

\newtheorem{theorem}{Theorem}[section]

\newtheorem{proposition}{Proposition}

\theoremstyle{definition}
\newtheorem{definition}[theorem]{Definition}

\DeclareMathOperator*{\argmin}{arg\,min}

\begin{document}
%
\title{Restoration of Atmospheric Turbulence-distorted Images via RPCA and Quasiconformal Maps}
%
%
%
%

\author{Chun Pong Lau, 
        Yu Hin Lai and
        Lok Ming Lui}
\markboth{Preprint}
{Shell \MakeLowercase{\textit{et al.}}: Bare Demo of IEEEtran.cls for Computer Society Journals}
%



\IEEEtitleabstractindextext{%
\begin{abstract}
We address the problem of restoring a high-quality image from an observed image sequence strongly distorted by atmospheric turbulence. A novel algorithm is proposed in this paper to reduce geometric distortion as well as space-and-time-varying blur due to strong turbulence. By considering a suitable energy functional, our algorithm first obtains a sharp reference image and a subsampled image sequence containing sharp and mildly distorted image frames with respect to the reference image. The subsampled image sequence is then stabilized by applying the Robust Principal Component Analysis (RPCA) on the deformation fields between image frames and warping the image frames by a quasiconformal map associated with the low-rank part of the deformation matrix. After image frames are registered to the reference image, the low-rank part of them are deblurred via a blind deconvolution, and the deblurred frames are then fused with the enhanced sparse part. Experiments have been carried out on both synthetic and real turbulence-distorted video. Results demonstrate that our method is effective in alleviating distortions and blur, restoring image details and enhancing visual quality.
\end{abstract}

}

\maketitle

\IEEEdisplaynontitleabstractindextext

%
\IEEEpeerreviewmaketitle

\IEEEraisesectionheading{\section{Introduction}\label{sec:introduction}}

\begin{figure}
\centering
 \includegraphics[width=0.4\textwidth]{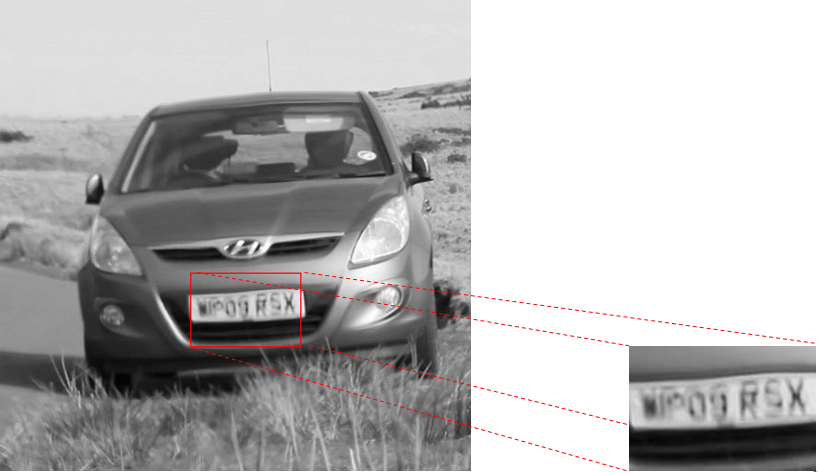}
 \caption{Left: The original high resolution video frame with mild distortion. Right: Zoomed object which has large deformation.}
 \label{figure:carfront_zoom}
\end{figure}

%
%
%
%
\IEEEPARstart{T}{he} problems of restoring a clear image from a sequence of turbulence-degraded frames are of high research interest, as the effect of geometric distortions and space-and-time-varying blur would significantly degrade image quality. Under the effects of the turbulent flow of air and changes in temperature, density of air particles, humidity and carbon dioxide level, the refractive index changes accordingly and light is refracted through several turbulence layers \cite{Roggemann1996} \cite{Hufnagel}. Therefore, when we want to capture images in locations where the temperature variation is large, for instance, deserts, roads with tons of vehicles, objects around flames, or from a long distance to perform long-range surveillance or to take pictures of the moon, rays from the objects would arrive at misaligned positions on the imaging plane, and thus distorted images are formed. Moreover, for a high-resolution video, even if the global oscillation of an image frame is not too large, the deformation of the objects in that image can be large. For example, in Figure \ref{figure:carfront_zoom}, the left image is an observed frame from a high-resolution mildly distorted video, while the right image is the same image zoomed in on a distorted object.\footnote{The image is from \url{http://seis.bris.ac.uk/~eexna/download.html}.} If we just consider the zoomed part in the video, the deformation would be large. In general, there are two types of approaches to deal with the problem, one being hardware-based adaptive optics techniques \cite{Pearson1976} \cite{Tyson1998} and the other being image-processing-based methods \cite{Shimizu} \cite{Dalong Li} \cite{evo eqn} \cite{filter flow} \cite{EFF}. In this paper, we focus on an image-processing-based method to restore the image. Since we are working on a sequence of distorted images or turbulence-degraded video, we assume the original image is static and the image sensor is also fixed. In order to model this problem, the mathematical model in this paper is based on  \cite{Zhu2013,Furhad2016},\begin{equation} \label{eq1}
I_t(x)=[D_t\big(H_t(I)\big)](x)+n_t(x),\indent t=1,\cdots,N
\end{equation}
where $I_t,I,$ and $n_t$ are the captured frame at time $t$, the true image, and the sensor noise respectively. The vector $x$ lies in the two-dimensional Euclidean space. $H_t$ represents the blurring operator, which is a space-invariant diffraction-limited point spread function (PSF). $D_t$ is the deformation operator, which is assumed to deform randomly. Note each of the sequences $\{D_t\}$ and $\{n_t\}$ are assumed to be identically distributed random variables, and the subscripts indicate the different actual outcomes that these variables turn out to be at different time instants.

Atmospheric turbulence has two main degradations on images: geometric distortion and blur. In this paper, we propose a new framework to stabilize a severely distorted video and reconstruct a sharp image with fine details.  First, we propose an iterative scheme to optimize an energy model to subsample sharp and mildly distorted video frames, and obtain a comparatively sharp reference image at the same time. This speeds up the computation and extracts the useful information from the original video. We then apply 2-step stabilization to stabilize the subsampled video with Beltrami coefficients, which further suppresses the distortion and replaces some comparatively blurry images with sharper ones by image warping using optical flow and RPCA. After that, we register the stabilized video with respect to the reference image by optical flow. Furthermore, we separate the video into a low-rank part and a sparse part by RPCA. On one hand, we apply blind deconvolution to deblur the low-rank part; on the other hand, we extract texture patches in the sparse part by adaptive thresholding, apply guided filtering to enhance the texture patches, and eventually fuse the low-rank and sparse parts together to obtain the final image.

\section{Previous work}
\label{section:Previous work}

Since the video frames are corrupted by both blur and geometric distortion, it is difficult to deal with them simultaneously, especially in the scenario where a large portion of the images are severely degraded. The registration process is furthered complicated by the lack of a good reference frame for the observed image sequence. Meinhardt-Llopis and Micheli \cite{Mario2014} proposed a reference extraction method which was coined the centroid method. In its scheme, the deformation fields between each pair of images are computed via optical flow and are assumed to have zero mean. This is equivalent to assuming the ground truth to be the temporal deformation mean of the frames. Next, every image is warped with the mean vector field to obtain a centroid image with respect to each image. Finally, the temporal mean of the centroids is taken to be the resultant image, in which the geometric deformations in the images are approximately cancelled out. In \cite{Mario2014linear}, Micheli {\it et al.} used a dynamic texture model to learn the parameters and put them into a Kalman filter model to get a blurry image. Finally, they used a Non-local Total Variation (NLTV) model to deblur and obtain a clear image. The assumption that the ground truth is the temporal deformation mean does not hold realistically. While the centroid method usually gives a good reference image, they cannot fully resolve the distortion, especially in the case that a large portion of the images are severely degraded. As the mean of the norms of deformation fields of the images increases, the deviation of the mean deformation of the images from the underlying model mean is amplified, and thus the resultant centroid images (and their temporal mean) remain distorted. Also, the temporal averaging makes the temporal mean of centroids contain features that appear in only a few frames, and thus ghost artifacts are formed.

Another method is the ``lucky frame" approach \cite{luckyframe}, which selects the sharpest frame from the video. This method is motivated by statistical proofs \cite{probability} that given sufficient video frames, there is a high probability that some frame would contain sharp texture details. Since in practice it is difficult to assume one can find a frame which is sharp everywhere, Aubailly {\it et al.} \cite{luckyregion} proposed the Lucky-Region method, which selects at each patch location the sharpest patch across the frames and fuses them together. Anantrasirichai {\it et al.} \cite{CLEAR2013} adopted this idea and introduced frame selection prior to registration. However, the cost function introduced was coarse, and the selection was done in one step by sorting. As a result, some of the selected frames geometrically differ significantly from the reference image. In addition, the cost function assumed the reference image (i.e. the temporal intensity mean over all frames) to accurately approximate the underlying true image, which is usually not the case. Another similar approach was proposed by Roggemann \cite{stat frame select}, where a subsample is selected from images produced by adaptive-optics systems to produce a temporal mean with higher signal-to-noise ratio.

As atmospheric turbulence can severely distort video frames, even if a satisfactory reference image is acquired, the video may not be registered well onto it. A feasible approach to enable registration is to stabilize the video and reduce the deformation between each frame and the reference image. Lou {\it et al.} \cite{Lou2013} proposed to stabilize video by sharpening each frame via spatial Sobolev gradient flow, and temporally smoothing the video to reduce interframe deformation. However, the distribution of the image intensities is not preserved under Sobolev gradient sharpening, and temporally smoothing produces ghost artifacts.

Zhu {\it et al.} \cite{Zhu2013} proposed a B-spline nonrigid registration algorithm to tackle distortion, and a patch-wise temporal kernel regression based near-diffraction-limited (NDL) image restoration to sharpen the image. Finally, they use blind deconvolution algorithm to deblur the fused image. However, NDL will further blur the image and produce some defects on the fused image. Then Furhad {\it et al.} proposed a frame selection criterion which is based on sharpness, in which they perform some preprocessing to filter out heavily blurred frames. They then propose spatiotemporal kernel regression to fuse the image. However, if sharp but severely distorted frames exist, they may be selected as deformation is not considered. Then the filtered image sequence will contain some distorted frames, which will degrade the reference image and hence the final output.

Recently, Robust Principal Component Analysis (RPCA) is another tool to tackle the problem of atmospheric turbulence. He {\it et al.} \cite{He2016} proposed a low-rank decomposition approach to separate the registered image sequence into low-rank and sparse parts. The former has less distortion, but is blurry and few detail texture image; on the other hand, the latter contains texture information but is noisy. Blind convolution is applied on the low-rank part to obtain a deblurred result, which is combined with the enhanced detail layer to get the final result. Xie {\it et al.} \cite{Xie2016} proposed a hybrid method, which assigns the low-rank image to be the initial reference image. The reference is then improved by solving a variational model, and the frames are registered to the reference image. However, as the deformation between the reference image and the observed frames may be large, direct registration may produce errors.  
\section{Contributions}
\label{section:Contributions}
\subsection{Frame sampling}
\label{subsection:Frame sampling}

In a turbulence-distorted video sequence of a stationary object and camera, all the frames are geometrically deformed and blurry. In order to remove the distortion by atmospheric turbulence, a good reference image is needed to obtain accurate deformation fields between the reference image and the observed frames. On the other hand, not all of the frames are useful to extract a good reference and provide useful information for image fusion. As a result, we propose subsampling the original video to obtain useful frames. The proposed algorithm can obtain a sharper and less deformed reference image and a good subsampled video by optimizing an energy model, which balances the number of subsampled frames and the quality of the reference image.

\subsection{Internal stabilization}
\label{subsection:Internal stabilization}
Suppose we have an image sequence in which a majority of frames are mildly deformed from each other, but the deformation of the remaining frames from the majority is significant. We propose extracting the low-rank part of the deformation fields among the frames via Robust Principal Component Analysis, in which the outlier deformation fields and error produced in the deformation estimation are diminished and reoriented. The frames are then warped with the adjusted fields, which aligns them well while avoiding heavy influence from outlier deformations.

\subsection{Absorbing stabilization}
\label{subsection:Absorbing stabilization}
Given an image sequence in which each frame is at most mildly deformed from each other, and additional frames which are deformed more severely, the latter can be stabilized by registering each of its frames to the aligned sequence. This allows for the incorporation of sharp but heavily deformed frames, which provides more texture details for image fusion afterwards.

The mathematical background and implementation details of the above methods will be elaborated in the following sections.

The remainder of this paper is organized as follows. The mathematical background of the employed conformal geometry techniques and Robust Principal Component Analysis is described in Section \ref{section:Mathematical Background}. The numerical scheme is described in detail in Section \ref{section:Proposed Algorithm}. The performance of the method is evaluated on six sets of videos/images and is compared with other techniques in Section \ref{section:Experimental Result and Discussion}. Finally, Section \ref{section:Conclusion} presents the conclusions of the paper.

\begin{figure*}[t]
\centering
\includegraphics[width=\textwidth]{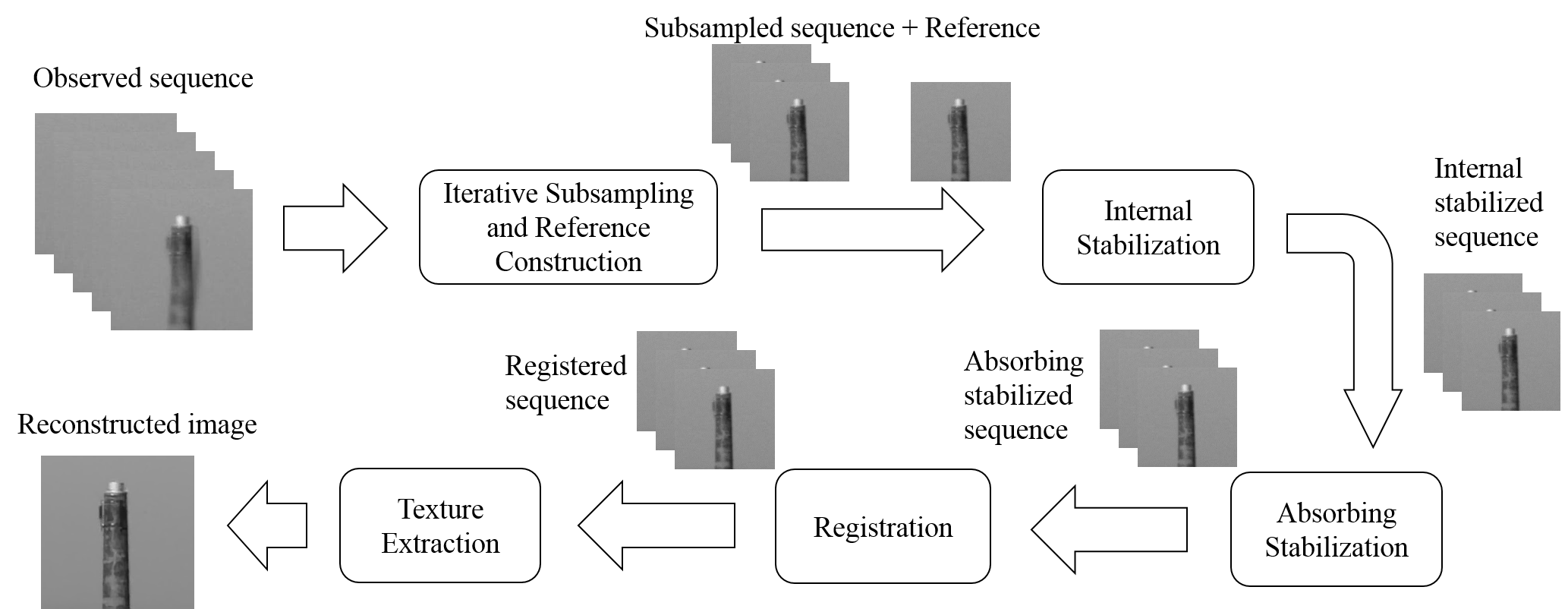}
\caption{Overall flow chart of the proposed algorithm.}
\label{flowchart_grid}
\end{figure*}

\section{Mathematical Background}
\label{section:Mathematical Background}
\subsection{Quasiconformal map}
\label{subsection:Quasi-conformal map}
\textit{Quasiconformal} maps are a generalization of conformal maps. They are orientation preserving
homeomorphisms between Riemann surfaces with bounded conformality distortion.
\begin{definition}
Let $f:\mathbb C\to\mathbb C$ be a continuous function with continuous partial derivatives $\dfrac{\partial f}{\partial z},\,\dfrac{\partial f}{\partial\bar z}$ defined via $\tilde f:\mathbb R^2\to\mathbb C$, $\tilde f(x,y):=f(x+iy)$, with $$\dfrac{\partial f}{\partial z}:=\dfrac{1}{2}\dfrac{\partial\tilde f}{\partial x}+\dfrac{1}{2i}\dfrac{\partial\tilde f}{\partial y},\indent\dfrac{\partial f}{\partial\bar z}:=\dfrac{1}{2}\dfrac{\partial\tilde f}{\partial x}-\dfrac{1}{2i}\dfrac{\partial\tilde f}{\partial y}.$$ $f$ is quasi-conformal provided that it satisfies the Beltrami equation,\begin{equation} \label{equation:beltrami}
\frac{\partial f}{\partial \bar{z}}=\mu(z)\frac{\partial f}{\partial z}
\end{equation}
 for some complex-valued Lebesgue measurable function $\mu$ satisfying $\Vert \mu \Vert_\infty < 1.$ $\mu$ is called the Beltrami coefficient (BC) of $f$.
\end{definition}
Given an orientation preserving homeomorphism $\phi$, we can find the corresponding BCs
from the Beltrami equation:\[ \mu_\phi=\frac{\partial \phi}{\partial \bar{z}}/\frac{\partial \phi}{\partial z} \] \\
The Jacobian $J$ of $\phi$ is related to $\mu_\phi$:\[ J(\phi)=\left\vert \frac{\partial \phi}{\partial z} \right\vert^2 \left(1-\vert\mu_\phi\vert^2 \right) \]\\
Since $\phi$ is an orientation preserving homeomorphism, $J(\phi) > 0$ and $\vert \mu_\phi\vert< 1$ everywhere. Hence, we must have $\Vert \mu_\phi \Vert_\infty\leq 1$. For any closed subset $D$ of $\mathbb C$, $\Vert\mu_{\phi|_D}\Vert_\infty<1$.
\begin{theorem}[measurable Riemann mapping theorem]

Suppose $\mu : \mathbb{C} \to \mathbb{C}$ is Lebesgue measurable and satisfies $\Vert \mu \Vert_\infty < 1$; then there is a quasiconformal homeomorphism $\varphi$ from $\mathbb{C}$ onto itself, which is in the Sobolev space $W^{1,2}(\mathbb{C})$ and satisfies the Beltrami equation in the distribution sense. Furthermore, by fixing 0, 1, and $\infty$, the associated quasiconformal homeomorphism $\varphi$ is uniquely determined.
\end{theorem}

Then a homeomorphism from $\mathbb{C}$ or $\mathbb{D}$ onto itself can be uniquely determined by its associated BC. Under this setting, if we are given a motion vector field between two frames, we can calculate the BC of the map. Conversely, we can also calculate the motion vector field if we have the associated BC.
\subsection{Linear Beltrami Solver}
\label{subsection:Linear Beltrami Solver}
Lui {\it et al}. \cite{Lui2013} proposed a linear algorithm, called Linear Beltrami Solver, to reconstruct a quasiconformal map $f$ from its associated Beltrami coefficient $\mu= \rho + i \tau$ on the rectangular domain $\Omega$ in $\mathbb{C}$. Let $f=u+\sqrt{-1}v$ , we have
\begin{multline} \label{equation:elliptic PDE}
\nabla \cdot  \bigg(A
\begin{pmatrix}
u_x \\
u_y
\end{pmatrix} \bigg)=0 , \nabla \cdot  \bigg(A
\begin{pmatrix}
v_x \\
v_y
\end{pmatrix} \bigg)=0, A=
\begin{pmatrix}
\alpha_1 & \alpha_2 \\
\alpha_2 & \alpha_3
\end{pmatrix};
\end{multline}
where 
$\begin{cases}
\alpha_1=\dfrac{(\rho-1)^2+\tau^2}{1-\rho^2-\tau^2}; \\
\alpha_2=-\dfrac{2\tau}{1-\rho^2-\tau^2}; \\
\alpha_3=\dfrac{(\rho+1)^2+\tau^2}{1-\rho^2-\tau^2},
\end{cases}$ and u and v satisfy some boundary conditions. In the discrete case, solving the above elliptic PDEs (\ref{equation:elliptic PDE}) can be discretized as solving a sparse symmetric positive definite linear system. Readers can refer to \cite{Lui2013} for details.

\subsection{Optical Flow}

Suppose we have a sequence of images, we can use optical flow to estimate the motion vector field between two images using the intensity difference between two images. Intuitively, optical flow is our visual sense of motion. Mathematically, for a three-dimensional case (with two spatial dimensions and a temporal dimension), a pixel at location $(x,y,t)$ with intensity $I(x,y,t)$ will be moved by $\Delta x, \Delta y$ and $\Delta t$ between the two image frames, and the following brightness constancy constraint can be given:
\begin{equation}
I(x,y,t)=I(x+\Delta x,y+\Delta y,t+\Delta t)
\end{equation}
By Taylor expansion, we have
\begin{multline}
I(x+\Delta x,y+\Delta y,t+\Delta t)\\
\approx I(x,y,t)+ \dfrac{\partial I}{\partial x} \Delta x+ \dfrac{\partial I}{\partial y} \Delta y+ \dfrac{\partial I}{\partial t} \Delta t
\end{multline}
So we have $I_xV_x+I_yV_y+I_t=0$. Thus, we have $\nabla I^T \cdot \overrightarrow{V}=-I_t$. This is an equation in two unknowns and cannot be solved as such. This is known as the aperture problem of the optical flow algorithms.

\subsubsection{Large Displacement Optical Flow}
In the paper, we will use Large Displacement Optical Flow to calculate the deformation fields between two image frames. We give a brief review on it.

Large Displacement Optical Flow \cite{optical flow 2009} is a coarse-to-fine variational framework for optical flow estimation between two image frames that incorporates descriptor matches in addition to the standard brightness and gradient constancy constraints. $w(x)=(u,v)$ is denoted as the displacement field between two images $I_1$ and $I_2$. Descriptor matches $w_{descr}(x)$ are obtained by matching densely sampled HOG descriptors in the two images with approximate nearest neighbour search.

$w(x)$ is obtained by minimizing the energy functional: \begin{align*}
E(\mathbf{w})&=\int_\Omega \Psi(|I_2(\mathbf{x}+\mathbf{w}(\mathbf{x}))-I_1(\mathbf{x})|^2)d\mathbf{x} \\
&+ \gamma \int_\Omega \Psi( |\nabla I_2(\mathbf{x}+\mathbf{w}(\mathbf{x}))-\nabla I_1 (\mathbf{x})|^2)d\mathbf{x} \\
&+ \alpha \int_\Omega \Psi(|\nabla u(\mathbf{x})|^2)+|\nabla v(\mathbf{x})|^2)d\mathbf{x} \\
&+\beta \int_\Omega \delta(\mathbf{x}) \rho(\mathbf{x}) \Psi(|\mathbf{w}(\mathbf{x})-\mathbf{w}_{descr}(\mathbf{x})|^2)d\mathbf{x}, \\
\Psi(s^2)&=\sqrt{s^2+10^{-6}}
\end{align*}

Here, $I_1$ and $I_2$ are the two input images, $\mathbf{w} := (u, v)$ is the
sought optical flow field, and $\mathbf{x} := (x, y)$ denotes a point
in the image.
In the above equation, the first two terms represent intensity and gradient constancy, the third term is the robust smoothness constraint and the last term biases the displacement field $w(x)$ towards the confident descriptor matches $w_{descr}(x)$. $\delta(x)$ is a delta function indicating whether a descriptor match is available in the location and $\rho(x)$ is the confidence of the match. Descriptor matches are obtained by matching densely sampled HOG descriptors in the two images. With the Large Displacement Optical Flow, we can calculate the motion vector field between two images even if the deformation between the two frames is relatively large. Readers can refer to \cite{optical flow 2009} for details.
\subsection{Low rank decomposition}
In mathematics, the low-rank approximation is a minimization problem, in which the cost function measures the fit between a given matrix (the data) and an approximating matrix (the optimization variable), subject to the constraint that the approximating matrix has reduced rank.

Suppose we have a matrix $R$. our goal is to decompose $R$ into $L+S$, where $L$ is the low-rank matrix and $S$ is the sparse matrix.
 We can obtain $R$ by \begin{equation}
\begin{cases}
\min\limits_{L,S} rank(L) + \lambda\Vert S \Vert_0  , \\
\text{ s.t. } L+S=R ,
\end{cases}
\end{equation}
where $\Vert E \Vert_0$ represents the number of non-zero entries in the matrix $E$.
Although the above problem follows naturally from our problem formulation, the cost function is highly non-convex and discontinuous, and the equality constraint is highly non-linear. Therefore, Candes et al. \cite{Candes} suggest that the above problem can be efficiently solved, under quite general conditions, by replacing the cost function,
\begin{equation}
\begin{cases}
\min\limits_{L,S} \Vert L \Vert_{\ast} + \lambda \Vert S \Vert_1  , \\
\text{ s.t. } L+S=R ,
\end{cases}
\end{equation}
where $\Vert E \Vert_{\ast}$ represents the nuclear norm of $E$ (the sum of its singular values), $\Vert E \Vert_1 $ represents the 1-norm of $E$ (the sum of absolute values of its entries), and $\lambda$ is a positive weighting factor. We call the above convex program Robust Principal Component Analysis (RPCA). For low-rank decomposition in this paper, we apply the algorithm of the Exact Augmented Lagrange Multiplier Method (EALM) \cite{EALM}, where the above convexified optimization problem is solved by Augmented Lagrange Multiplier method with the augmented problem $(L,S)=\arg\min\limits_{A,E}\Vert A\Vert_*+\lambda_\text{RPCA}\Vert E\Vert_1+\langle Y,R-A-E\rangle+\dfrac{\mu}{2}\Vert R-A-E\Vert_F^2$.

\begin{figure}
\centering
 \includegraphics[width=0.3\textwidth]{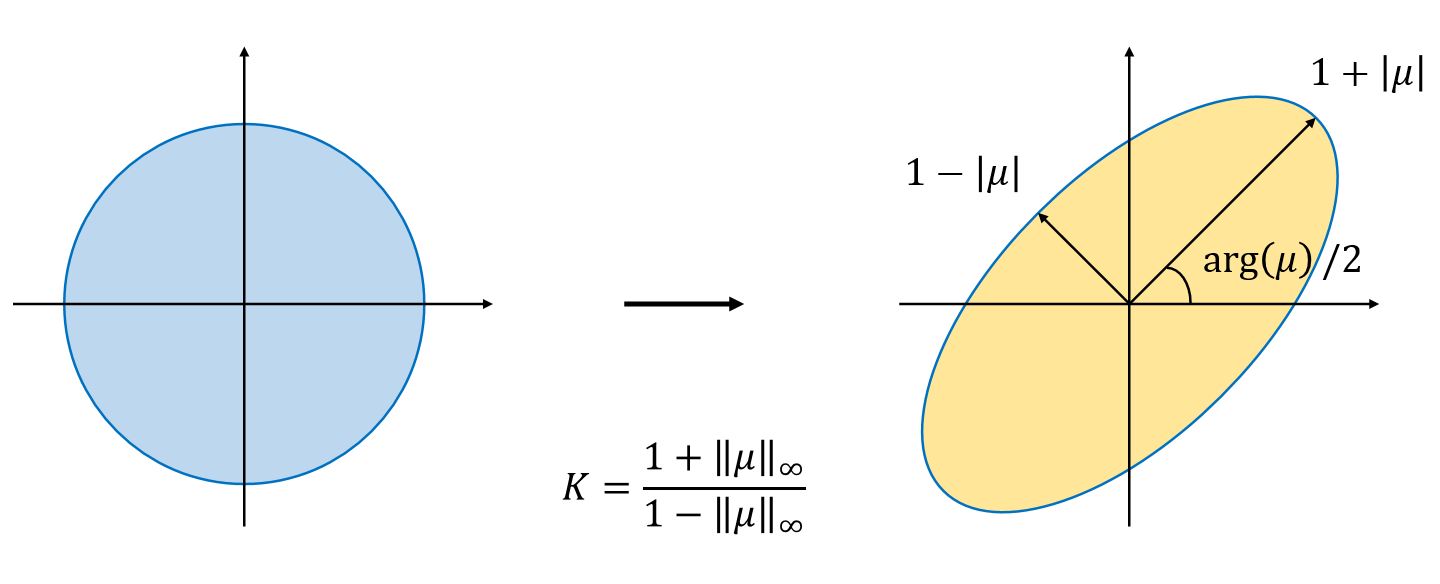}
 \caption{An illustration of quasi-conformal mappings. The maximal magnification and shrinkage as well as the local rotational angle are determined by the Beltrami coefficient $\mu$ of the mappings.}
 \label{figure:beltrami}
\end{figure}

\begin{figure}
\centering
 \includegraphics[width=0.4\textwidth]{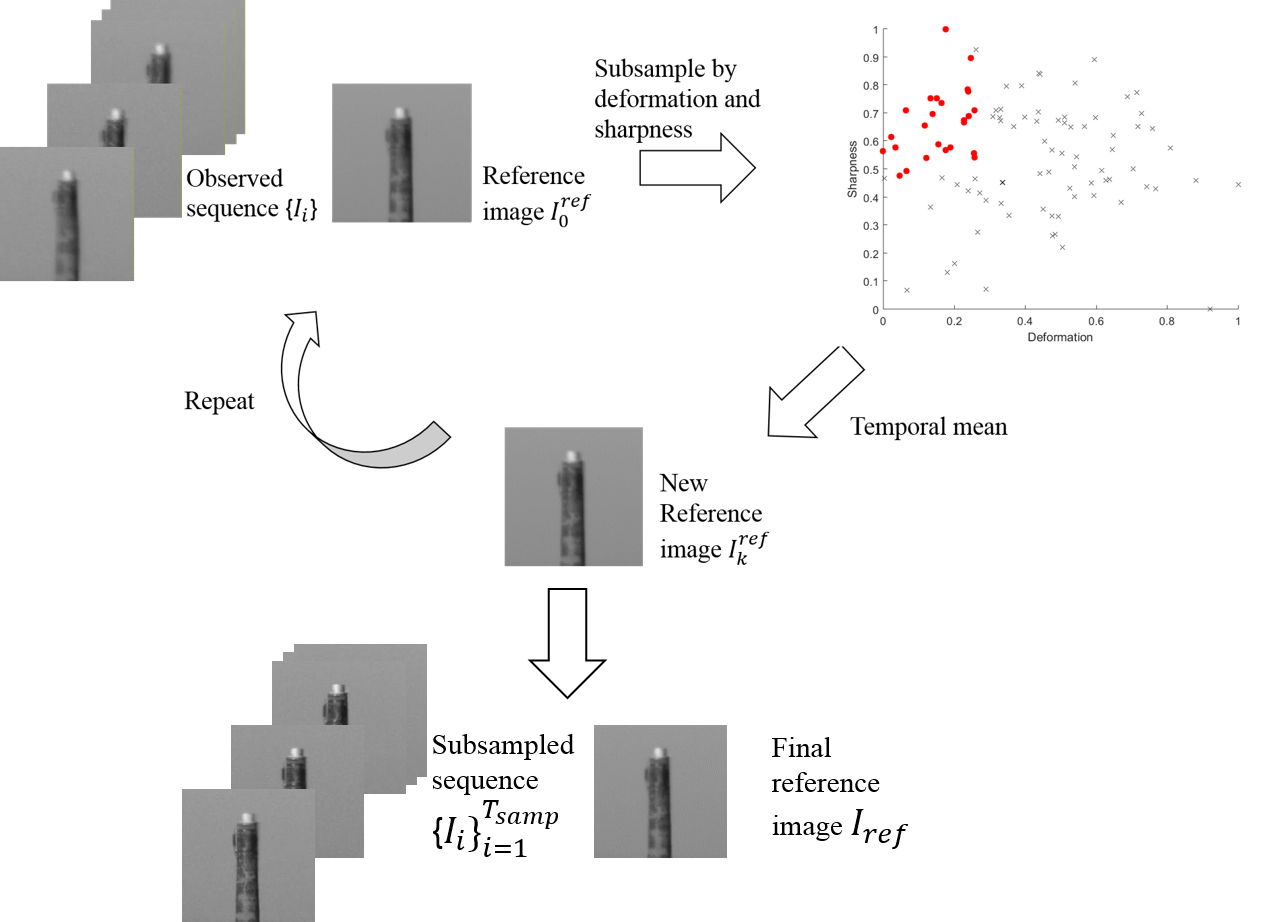}
 \caption{Block diagram for optimizing $E_Q(I,J)$.}
 \label{figure:sub-sample}
\end{figure}

\section{Proposed Algorithm}
\label{section:Proposed Algorithm}
In this section, we describe our proposed algorithm in detail. Our algorithm can be divided into three main stages (see Figure \ref{flowchart_grid}), namely,
1. Reference image extraction and subsampling, 2. Stabilization and 3. Image fusion.

\subsection{Reference image extraction and subsampling}
\label{subsection:Reference image extraction and sub-sampling}

To restore a turbulence-degraded video, a suitable reference image capturing the geometric structures of objects in the image frames is crucial. With a good reference image, the geometric deformation under turbulence can be accurately estimated through image registration between the reference image and image frames. The extraction of the reference image certainly relies on the image frames of the video. However, under turbulence distortions, not all image frames are useful for extracting a reference image. For example, image frames with large geometric deformations and blurs will not provide accurate information to extract the reference image. Therefore, it calls for developing an algorithm to simultaneously subsample useful frames from the video and extract the reference image from the subsampled frames.

We propose an iterative algorithm to subsample frames of the video and extract an accurate reference image by considering an optimization problem. Useful image frames should be sharp and less distorted. On the other hand, if more frames that are useful are subsampled, more information can be used to extract the reference image. Therefore, it is necessary to design an algorithm that optimizes between the number of frames and the usefulness of these subsampled frames.

\begin{figure}[t]

\centering
\begin{subfigure}[t]{0.15\textwidth}
\includegraphics[width=\textwidth]{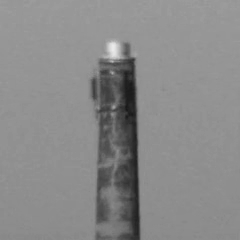}
\caption{Sharp observed image in Chimney sequence}
\end{subfigure}\quad
\begin{subfigure}[t]{0.15\textwidth}
\includegraphics[width=\textwidth]{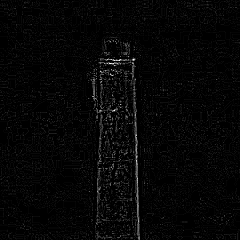}
\caption{Laplacian image of (A)}
\end{subfigure}\quad \\
\begin{subfigure}[t]{0.15\textwidth}
\includegraphics[width=\textwidth]{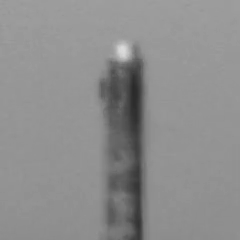}
\caption{Blurry observed image in Chimney sequence}
\end{subfigure}\quad
\begin{subfigure}[t]{0.15\textwidth}
\includegraphics[width=\textwidth]{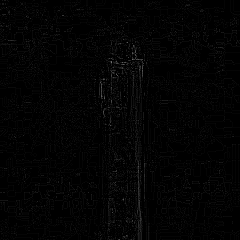}
\caption{Laplacian image of (C)}
\end{subfigure}\quad
\caption{Illustration of Sharpness indicator.}
\label{figure:Sharpness}
\end{figure}

Denote the image frames of a turbulence-degraded video by $\{I_t=I_t(x,y)\}_{t=1}^T$ , where $(x,y)$ denotes a pixel in the image domain $\Omega$. Our goal is to search for an optimal subsample of indices $J\subset \{1,2,\dots,T\}$, such that a good reference image $I^{ref}$ can be extracted from $\{I_i\}_{i\in J}$. To achieve this goal, we first need to quantitatively measure the sharpness and geometric distortion of an image frame. The sharpness $\mathcal{S}(I)$ of an image $I$ can be measured by
\begin{equation}
    \mathcal{S}(I)=\Vert \Delta I \Vert_1
\end{equation}
\noindent where $\Delta$ is the Laplacian operator. In essence, $\Delta I$ is the convolution of $I$ with the Laplacian kernel, which captures the features or edges of objects in the image (see Figure \ref{figure:Sharpness}). The magnitude of $\Delta I$ is higher for sharper images.  Hence, $\mathcal{S}(I)$ is larger for sharper images. For example,  Figure \ref{figure:Sharpness}(b) and (d) show the Laplacian image of a sharp image (Figure \ref{figure:Sharpness}(a)) and a blurry image (Figure \ref{figure:Sharpness}(c)) respectively. The magnitude in intensities of Figure \ref{figure:Sharpness}(c) is obviously larger than that of Figure \ref{figure:Sharpness}(d). Hence, $\mathcal{S}(I)$ provides an effective measurement for the sharpness of an image. For the ease of implementation, we usually normalize $\mathcal{S}(I)$ to the range of $[0,1]$.

Next, we need to quantitatively measure the geometric distortion. An intuitive way to measure the geometric distortion between two images is to estimate the deformation field between them. However, it involves high computational costs to compute the image correspondences. To alleviate this issue, we propose to measure the geometric distortion between two images by measuring their dissimilarities in intensities at every pixel. If two images are similar in intensities, it requires less deformation to transform one image to another. Common measures of dissimilarity include the sum of squared differences $\Vert I^\text{ref}-I_i\Vert_2^2$, the sum of absolute differences $\Vert I^\text{ref}-I_i\Vert_1$ and the weighted sum of absolute and gradient differences $\Vert I^\text{ref}-I_i\Vert_1+\gamma\Vert \nabla I^\text{ref}-\nabla I_i\Vert_1$. In this paper, we adopt the sum of squared differences as the dissimilarity measure.

\begin{figure*}[t]
\centering
\includegraphics[width=0.9\textwidth]{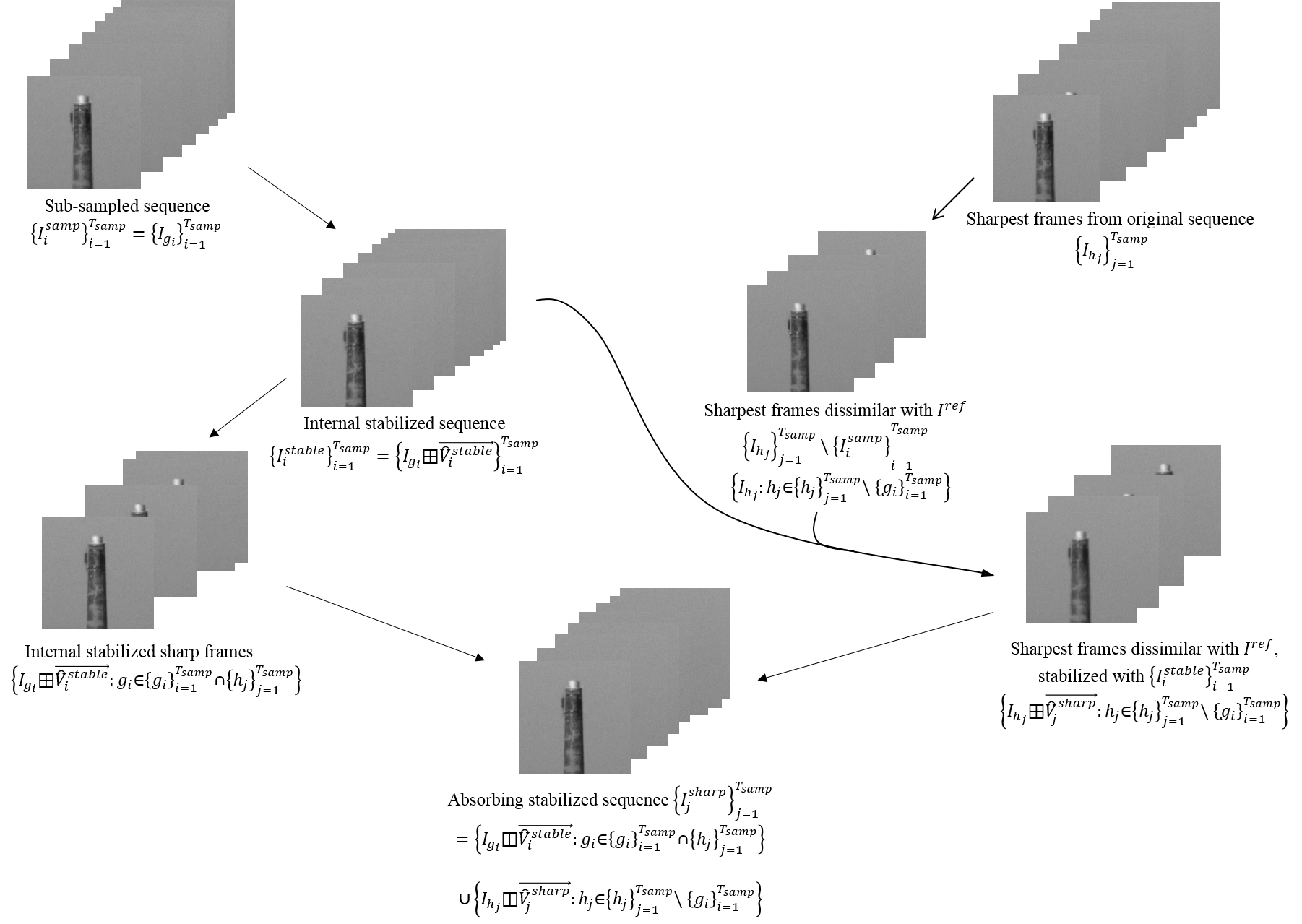}
\caption{Block diagram for Stabilization.}
\label{figure:stabilization}
\end{figure*}

Now, in order to obtain an optimal subsample $\{I_i\}_{i\in J^*}$ of the image frames as well as the optimal reference image $I^\text{ref}$, we propose to maximize the following energy functional:
\begin{equation}\label{optimizationmodel}
E(I,J) = \alpha (1-e^{-\rho |J|})-E_{Q}(I,J),
\end{equation}
\noindent where $E_Q(I,J)$ measures the sharpness and the geometric distortion from the potential reference image $I$ of the subsampled frames $\{I_i\}_{i\in J}$. The term $E_Q(I,J)$ will be defined later. Here, $\alpha >0$ is a parameter balancing the first and second terms in the energy functional. The first term is an increasing function in the cardinality $|J|$ of $J$, which aims to subsample as many useful frames as possible for the extraction of the reference image. Note that a concave increasing function is chosen, as a marginal increase in the size of the subsample has reduced effect on the accuracy of the extracted reference image as the number of subsampled frames increases. Therefore, by maximizing the energy functional with the two combined terms, our model simultaneously searches for a good reference image together with a maximal subsample of sharp frames, whose geometric distortions from the reference image are small.  

Next, we proceed to define $E_Q(I,J)$. Using the measures of sharpness and geometric distortion described above, $E_Q(I,J)$ can be formulated as follows: 
\begin{equation}
    E_{Q}(I,J)=\dfrac{1}{|J|}  \sum\limits_{i\in J} [\Vert I-I_i \Vert_2^2+\lambda_\text{samp}(1-\mathcal{S}(I_i))],
\end{equation}
where $\lambda_\text{samp} >0$ is a positive constant for controlling the importance of sharpness of the image frames. The first term measures the geometric distortion of $I_i$ from $I$. Clearly, $E_Q(I,J)$ is small if the subsampled frames are sharp and their geometric distortions from $I$ are small.

To increase the energy functional $E(I,J)$, the following strategy can be applied. Fixing $|J|= k$, we consider the following optimization problem:
\begin{equation}
\label{Energy}
    \begin{cases}
    (I_k^\text{ref},J^k)=\mathbf{argmin}_{I,J} E_Q(I,J), \\
    \text{ such that } |J|=k.
    \end{cases}
\end{equation}

Suppose $k^* = \mathbf{argmax}_k  \alpha (1-e^{-\rho k})-E_{Q}(I_k^\text{ref},J_k)$. Then, $(I_{k^*}^\text{ref},J^{k^*})$ solves the optimization problem (\ref{optimizationmodel}). In practice, we compute the finite sequence $\{ \mathcal{E}_k:=\alpha (1-e^{-\rho k})-E_{Q}(I_k,J_k)\}_{k=1}^T$ and pick the largest $\mathcal{E}_{k^*}$. Then, $J^* = J^{k^*}$ and $I^\text{ref} = I_{k^*}^\text{ref}$ solve our proposed optimization model.

Now, to solve the optimization problem (\ref{Energy}), an alternating minimization scheme is applied. Supposed $\lambda_\text{samp}$ and $k$ are fixed, and an initial subsampling $J^0$ with $|J^0|=k$ is arbitrary chosen. The iterative scheme can then be described as follows: 
\begin{enumerate}
    
    \item Fixing $I=I^{\text{iter}-1}$, we minimize $E_Q$ over $J$. Note that  $\Vert I-I_i \Vert_2^2$ and $\lambda_\text{samp}(1-\mathcal{S}(I_i))$ can both be easily calculated for each $i$. Denote $\Vert I-I_i \Vert_2+\lambda_\text{samp}(1-\mathcal{S}(I_i))$ by $E_i$. Arrange $E_i$ such that:
    \begin{equation}
        E_{i_1}\leq E_{i_2}\leq ...\leq E_{i_j}\leq ... \leq E_{i_T}.
    \end{equation}
    Then, $J^* = \{i_1,i_2,...,i_k\}$ is the required minimizer. We set $J^\text{iter} = J^*$ and compute $E_Q^\text{iter}=\frac{1}{k}\sum\limits_{i\in J^*}E_i$.
    
    \item Next, fixing $J=J^{\text{iter}}$, we minimize $E_Q$ over $I$. Note that the second term $\frac{1}{k}\sum\limits_{i\in J}\lambda_\text{samp}(1-\mathcal{S}(I_i))$ is a constant since $J$ is fixed. We only need to find $I$ that minimizes $\dfrac{1}{k} \sum\limits_{i\in J} \Vert I-I_i \Vert_2^2$. By differentiating with respect to $I(x,y)$, the minimizer is given by the temporal mean $\bar{I}_J$ of $\{I_i\}_{i\in J}$:
    \begin{equation}
        \bar{I}_{J^\text{iter}} =\frac{1}{k}\sum_{i\in J^\text{iter}} I_i. 
    \end{equation}
    We then set $I^\text{iter} = \bar{I}_{J^\text{iter}}$.

\end{enumerate}
Repeat step 1 and step 2 above until the difference $ DE_Q=E_Q^{\text{iter}-1}-E_Q^\text{iter}$ between the energies at the current and previous steps is smaller than some hyperparameter $\varepsilon$. Note that $E_Q^\text{iter}$ is always lower than $E_Q^{\text{iter}-1}$, as each step in the alternating scheme forces the energy to drop or remain the same. Figure \ref{figure:sub-sample} illustrates the optimization of $E_Q(I,J)$.

The overall algorithm is summarized in Algorithm \ref{alg:Reference extraction}.

\begin{algorithm}[t]
\begin{algorithmic}[1]
\REQUIRE Video sequence $\{I_t=I_t(x,y):(x,y) \in \Omega\}_{t=1}^T$
\ENSURE Subsampled image sequence $\{ I_i\}_{i=1}^{T_\text{samp}}$ with less distortion and higher sharpness; Reference image $I^\text{ref}(x,y)$
\STATE Compute the sharpness $\mathcal S(I_i)$ of each frame $\{I_i\}_{i=1}^T$

\FOR{$k=2$ to $T$}
\STATE Obtain the initial reference image $I^0=\bar{I}_{J^0}$, where $J^0$ with $|J^0|=k$ is arbitrarily chosen 

\WHILE{$DE_Q>\varepsilon$}
\STATE Calculate $\Vert I^{\text{iter}-1}-I_i \Vert_2^2$ for each $i$
\STATE Calculate $E_Q^\text{iter}$ by considering $J$ which contains $\{ I_i \}$ with the k smallest $\Vert I-I_i \Vert_2^2+\lambda_\text{samp}(1-\mathcal S(I_i))$
\STATE $I^\text{iter} = \bar{I}_J$ 
\STATE Calculate the difference $DE_Q=E_Q^{\text{iter}-1}-E_Q^\text{iter}$ between the current and previous subsamples

\ENDWHILE
\STATE Calculate the total energy $E_k=\alpha(1-e^{-\rho k})-E_Q^\text{final}$
\ENDFOR
\RETURN the subsampled sequence $J^\ast$ with the maximum total energy and reconstruct the final reference image $I^\text{ref}=\bar{I}_{J^\ast}$
\end{algorithmic}
\caption{Subsampling and reference extraction}
\label{alg:Reference extraction}
\end{algorithm}

\subsection{Stabilization}\label{subsec:Stabilization}
After the video is subsampled, sharp image frames with comparatively smaller geometric distortions remain.
The subsampled frames can further be stabilized by warping each image frame via a suitable deformation field. As a result, a significant amount of geometric deformations can be removed. The stabilization involves two procedures (see Figure \ref{figure:stabilization}), namely, (1) Internal stabilization and (2) Absorbing stabilization. We will describe each procedure in detail.

\begin{figure}[t]

\centering
\begin{subfigure}[t]{0.22\textwidth}
\includegraphics[width=\textwidth]{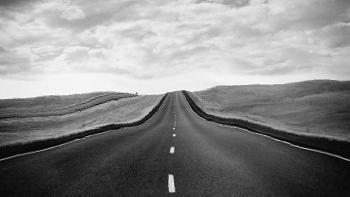}
\vspace{-1em}
\end{subfigure}
\begin{subfigure}[t]{0.22\textwidth}
\includegraphics[width=\textwidth]{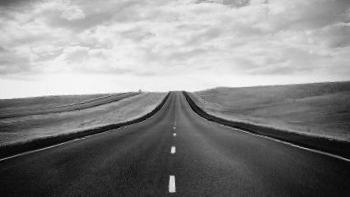}
\vspace{-1em}
\end{subfigure}\\ 

\begin{subfigure}[t]{0.22\textwidth}
\includegraphics[width=\textwidth]{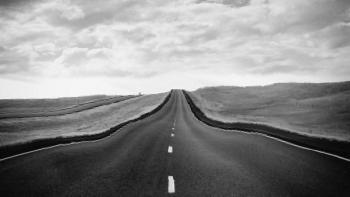}
\caption{}
\end{subfigure} 
\begin{subfigure}[t]{0.22\textwidth}
\includegraphics[width=\textwidth]{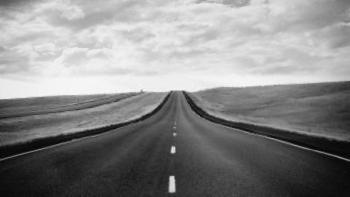}
\caption{}
\end{subfigure}
\caption{Stabilizing the comparatively severely deformed frames in the Road sequence. (a) Comparatively severely deformed frame in subsampled sequence. (b) Stabilized.}
\label{figure:road stable}
\end{figure}

\subsubsection{Internal stabilization}\label{subsubsec:InternalStabilization}
Although a subsample of frames has been chosen in the previous step by minimizing geometric deformation, one may still not assume the subsampled frames are free of deformation and blur, if most frames of the original sequence are severely distorted. As a result, they cannot be registered properly. Unsatisfactory registration will seriously influence the fusion of the images. In order to reduce the error of registration, we propose a stabilization method which further suppresses the distortion. See Figure \ref{figure:road stable}.

Let $\{ I_j^{\text{samp}} \}_{j=1}^{T_{\text{samp}}}$ be the subsampled frames obtained from the previous step. Our goal is to obtain a new image sequence $\{ I_j^{\text{stable}} \}_{j=1}^{T_{\text{samp}}}$ from $\{ I_j^{\text{samp}} \}_{j=1}^{T_{\text{samp}}}$ with less oscillation, such that each new frame is closer to the ground truth image $I$. 

Suppose the reference image $I^{\text{ref}}$ from the previous step is close to $I$. Then, an intuitive way to stabilize $\{ I_j^{\text{samp}} \}_{j=1}^{T_{\text{samp}}}$ can be done by warping each frame $I_j^{\text{samp}}$ by the registration map $D_{\text{ref}}^j : \Omega \rightarrow \Omega$ from $I^{\text{ref}}$ to $I_j^{\text{samp}} $.
More specifically, we obtain $I_j^{\text{stable}}$ by $$I_j^{\text{stable}}=I_j^{\text{samp}} \circ D_{\text{ref}}^j. 
$$ This approach to stabilize the video is effective if the reference image and video frames are both sharp and mildly distorted. In our case, since the video is severely distorted by turbulence, both the extracted reference image and the subsampled frames may suffer from various levels of blur and geometric distortion. Then objects in the reference image may be too blurry or occluded for optical flow algorithms to establish correct correspondences. As a result, this simple method to stabilize the video by registering each frame to the reference image is not applicable. To alleviate this issue, a remedy is to consider the deformation map $f_i^j : \Omega \rightarrow \Omega$
from $I_i^{\text{samp}}$ to $I_j^{\text{samp}}$. It is much less likely that a particular object is blurred or occluded in each of the frames $I_j$.
By incorporating data from all of $I_j^\text{samp}$, we can avoid the aforementioned errors while reducing geometric distortions.

Our goal is to extract an accurate estimation of the deformation map $D^{\text{ref}}_i : \Omega \rightarrow \Omega$
from each frame $I_i^{\text{samp}}$  to the reference image. The registration map $D^{\text{ref}}_i$ can be represented by the deformation field $\overrightarrow{V^{\text{ref}}_i}$, where 
$$D^{\text{ref}}_i(p)=p+\overrightarrow{V^{\text{ref}}_i}(p) \quad  \forall p \in \Omega.$$ Similarly $f_i^j$ can be represented by a vector field $\overrightarrow{V_i^j}$, where $$f_i^j(p)=p+\overrightarrow{V_i^j}(p) \quad  \forall p \in \Omega.$$ Since the reference image is the temporal mean of the subsampled frames, the temporal mean of the deformation fields $\{\overrightarrow{V^i_{\text{ref}}}\}$ is usually small. Mathematically, one can assume that $$\Bigg\Vert\dfrac{\sum\limits_{i=1}^{T_\text{samp}} \overrightarrow{V^i_{\text{ref}}}}{T_\text{samp}}\Bigg\Vert_{\infty} \leq \varepsilon \quad \text{ for small } \varepsilon > 0.$$ Then the following proposition is a useful observation for the extraction of $D^{\text{ref}}_i$ from $\{ f_i^j \}_{j=1}^{T_\text{samp}}$. 

\begin{proposition}\label{proposition:field_error}
Suppose $\Bigg\Vert\dfrac{\sum\limits_{i=1}^{T_\text{samp}} \overrightarrow{V^i_\text{ref}}}{T_\text{samp}}\Bigg\Vert_{\infty} \leq \varepsilon$. Let $\tilde{D_i}(p)= p+ \dfrac{\sum\limits_{j=1}^{T_\text{samp}} \overrightarrow{V_i^j}(p)}{T_\text{samp}} $. Then $\tilde{D_i} \circ D^i_{\text{ref}} \approx \textbf{id}$, i.e. $\Vert\tilde D_i\circ D^i_\text{ref}-\textbf{id}\Vert_\infty\leq\varepsilon$.

\begin{proof}
\begin{align*}
f_i^j \circ D^i_{\text{ref}}(p)&=f_i^j(p+\overrightarrow{V^i_{\text{ref}}}(p)) \\ &=p+\overrightarrow{V^i_{\text{ref}}}(p)+\overrightarrow{V_i^j}(p+\overrightarrow{V^i_{\text{ref}}}(p));\\
D^j_{\text{ref}}(p)&=f_i^j\circ D^i_{\text{ref}}(p)\\
\text{By averaging over $j$,}\\
{\implies} p+ \dfrac{\sum\limits_{j=1}^{T_\text{samp}} \overrightarrow{V^j_{\text{ref}}}(p)}{T_\text{samp}}
&= p+\overrightarrow{V^i_{\text{ref}}}(p)+ \dfrac{\sum\limits_{j=1}^{T_\text{samp}} \overrightarrow{V_i^j}(p+\overrightarrow{V^i_{\text{ref}}}(p))}{T_\text{samp}} \\
\implies\tilde{D_i} \circ D^i_{\text{ref}}(p)-p&=\dfrac{\sum\limits_{j=1}^{T_\text{samp}}\overrightarrow{V_\text{ref}^j}(p)}{T_\text{samp}}\\
\implies\Vert\tilde D_i\circ D_\text{ref}^i-\textbf{id}\Vert_\infty&\leq\varepsilon.
\end{align*}
\end{proof}
\end{proposition}

\begin{figure}[t]

\centering
\begin{subfigure}[t]{0.15\textwidth}
\includegraphics[width=\textwidth]{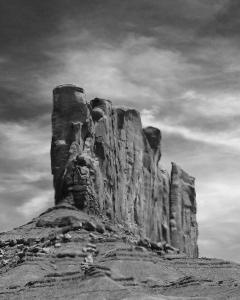}
\caption{Mildly distorted image before stabilized}
\end{subfigure}
\begin{subfigure}[t]{0.15\textwidth}
\includegraphics[width=\textwidth]{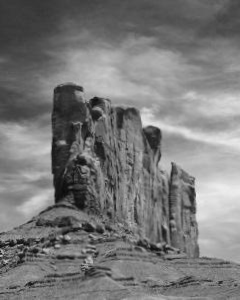}
\caption{Centroid method, PSNR=25.2853}
\end{subfigure}
\begin{subfigure}[t]{0.15\textwidth}
\includegraphics[width=\textwidth]{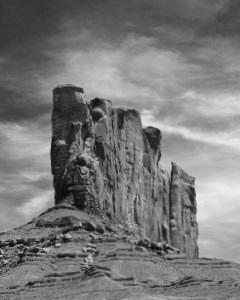}
\caption{Internal stabilization, PSNR=27.4840}
\end{subfigure}
\caption{Comparison of small deformed image stabilized in Centroid method and Internal stabilization in Desert sequence. \iffalse The video consists of 30\% mildly distorted frames and 70\% heavily distorted frames.\fi}
\label{figure:desert stable}
\end{figure}

This proposition gives guidance on how one can estimate $D^{\text{ref}}_i$ with $\tilde D_i$ from $\{ f_i^j \}_{j=1}^{T_{\text{samp}}}$. Pick a pixel $p$ of the reference image $I^{\text{ref}}$. Due to the turbulence, the position of $p$ is warped to $D^i_{\text{ref}}(p)$ in $\{ I_i^{\text{samp}} \}_{i=1}^{T_{\text{samp}}}$ causing geometric deformation in the image frames. To remove the geometric deformation, one can warp the point $D^i_{\text{ref}}(p)$ by $\tilde{D_i}$ to a new position $\tilde{D_i}(D^i_{\text{ref}}(p))$. According to the proposition, $\tilde{D_i}(D^i_{\text{ref}}(p))$ is close to the original position of $p$ in the reference image $I^{\text{ref}}$. Hence, the geometric deformation can be suppressed. 

On the other hand, by triangle inequality, it is easily to observe that $$\Vert \tilde{D_i}(D^i_{\text{ref}}(p))-\tilde{D_j}(D^j_{\text{ref}}(p)) \Vert_{\infty} \leq 2\varepsilon\text{ for any } i \neq j$$ Thus, the new image sequence $\{ I_i^{\text{stable}} := I_i^{\text{samp}} \boxplus \overrightarrow{\tilde{V_i}}\}_{i=1}^{T_{\text{samp}}}$ after warping has small oscillation provided that $\varepsilon$ is small, where $\boxplus$ is the warping operator and $\overrightarrow{\tilde V_i}(p):=\tilde D_i(p)-p$. 

This method has also been applied in \cite{Mario2014},\cite{Mario2012}, and has shown itself to be effective for suppressing geometric distortion, provided that the deformation field is small and the video is not too blurry.


\begin{algorithm}[t]
\begin{algorithmic}[1]
\REQUIRE Subsampled image sequence $\{ I_i^\text{samp}\}_{i=1}^{T_\text{samp}}$
\ENSURE Stabilized image sequence $\{ I_i^{\text{stable}} \}_{i=1}^{T_\text{samp}}$
\FOR{$i=1$ to $T_\text{samp}$}
\FOR{$j \neq i$}
\STATE Compute the deformation field $\overrightarrow{V_i^j}$
\ENDFOR
\STATE Apply RPCA on $\mathcal V_i$ to obtain $\{\mathcal L_{i,p}:p=1,2\}$
\STATE Calculate Beltrami representation $\mu_i^\text{stable}$ of $\overrightarrow{V_i^\text{stable}}$
\STATE Restrict $\Vert \hat\mu_i^\text{stable} \Vert_{\infty} < 1$
\STATE Reconstruct the fold-free deformation field $\overrightarrow{\hat{V}_i^\text{stable}}$ by applying LBS  
\STATE Obtain $I_i^{\text{stable}}$ by warping $I_i$ with $\overrightarrow{\hat{V}_i^\text{stable}}$  
\ENDFOR
\RETURN $\{ I_i^{\text{stable}} \}_{i=1}^{T_\text{samp}}$
\end{algorithmic}
\caption{Internal stabilization}
\label{algorithm:Internal stabilization}
\end{algorithm}

However, in some cases, a few of the image frames $\{I_i^\text{samp}\}$ will contain deformations that differ significantly from the remaining majority of frames. These outlying deformed frames may originate from incorrect registration via optical flow, or from subsampled frames which have relatively larger displacement from the reference image. Since the centroid method simply calculates the mean of the motion vector fields, if there are some outlier deformations, the centroid method will also capture those deformations and hence the shape of those small deformed frames will be deformed. To solve the problem, we propose to apply RPCA on the deformation fields to suppress the outlier deformations. See Figure \ref{figure:desert stable}.

For every frame $I^\text{samp}_i$, we calculate the deformation fields $\overrightarrow{V_i^j}$ from the fixed frame $I^\text{samp}_i$ to the other frames. Denote vectorized $f$ as $\text{vec}(f)$. 
Define \begin{multline*}\mathcal V_i:=\begin{pmatrix}|&|& &|\\
 \text{vec}(\overrightarrow{V_i^1})&\text{vec}(\overrightarrow{V_i^2})&\cdots&\text{vec}(\overrightarrow{V_i^{T_\text{samp}}})\\
 |&|& &|\end{pmatrix}\\
=\text{Re}(\mathcal V_i)+i\,\text{Im}(\mathcal V_i)\end{multline*}$=\mathcal V_{i,1}+i\mathcal V_{i,2}$, where $\mathcal V_{i,1}:=\text{Re}(\mathcal V_i)$ contains the horizontal displacement vectors, and $\mathcal V_{i,2}:=\text{Im}(\mathcal V_i)$ contains the vertical displacement vectors.

We then apply RPCA to decompose each of $\{\mathcal V_{i,p}:p=1,2\}$ into low-rank and sparse parts:
 \begin{align*}
\mathcal V_{i,p}&=\mathcal L_{i,p}^*+\mathcal S_{i,p}^*\\
(\mathcal L_{i,p}^*,\mathcal S_{i,p}^*)&=\argmin\limits_{\substack{\mathcal L,\mathcal S\\
\mathcal L+\mathcal S=\mathcal V_{i,p}}}\Vert\mathcal L\Vert_*+\lambda\Vert\mathcal S\Vert_1\text{ for }p=1,2
\end{align*}
 Denote $$\mathcal L_{i,p}^*=\begin{pmatrix}|&|& &|\\
 \text{vec}(\overrightarrow{L_{i,p}^1})&\text{vec}(\overrightarrow{L_{i,p}^2})&\cdots&\text{vec}(\overrightarrow{L_{i,p}^{T_\text{samp}}})\\
 |&|& &|\end{pmatrix},$$ where $\overrightarrow{L_{i,p}^j}$ are deformation fields each of size $X\times Y$.
 By the above argument, in the presence of outlier deformation fields, we obtain the stabilized deformation field $\overrightarrow{V_i^\text{stable}}$ in the mean of $\{\overrightarrow{L_{i,p}^j}\}_{j=1}^{T_\text{samp}}$. When the deformation fields exhibit an overall pattern, the RPCA algorithm extracts the part that resembles the other fields in the low-rank part, whereas the outlying part is captured in the sparse matrix. However, when there is no such pattern, then the low-rank part extracted by RPCA will not correspond to any useful pattern. We identify the absence of a general pattern by observing the number of non-zero entries in the sparse part by RPCA. If too many non-zero entries are present, we resort to applying centroid method.
$$\overrightarrow{V_i^\text{stable}}=\begin{cases}\dfrac{1}{T_\text{samp}} \sum\limits_{j=1}^{T_\text{samp}}\overrightarrow{L_{i,p}^j}&\text{if }\Vert\mathcal S_{i,p}^*\Vert_0<\dfrac{XYT_\text{samp}}{2}\\
\dfrac{1}{T_\text{samp}} \sum\limits_{j=1}^{T_\text{samp}}\overrightarrow{V_{i,p}^j}&\text{otherwise}\end{cases}.$$

On the other hand, warping an image with a non-bijective deformation field may often cause unnatural artifacts. Since the object and the camera are static, we can require that the deformation field is bijective. We can easily enforce the bijectivity constraint by smoothing out the Beltrami coefficient, and guarantee the warped image is fold-free. By Beltrami equation (\ref{equation:beltrami}), we can calculate $\mu_i^\text{stable}$, the Beltrami representation of $\overrightarrow{V_i^\text{stable}}$ for each $i$. Then we restrict $\Vert \mu_i^\text{stable} \Vert_{\infty} < 1$ by thresholding it, i.e. $$\hat\mu_i^\text{stable}=\begin{cases}\mu_i^\text{stable}&\text{if }|\mu_i^\text{stable}|<1\\
\dfrac{\mu_i^\text{stable}}{|\mu_i^\text{stable}|+\varepsilon}&\text{otherwise}\end{cases}.$$ As seen in Figure \ref{figure:beltrami}, at each position where the Beltrami representation is thresholded, the orientation (i.e. arg($\mu$)/2) of local deformation is preserved.

Then we can reconstruct the deformation field with the Linear Beltrami Solver: $$\overrightarrow{\hat{V}_i^\text{stable}} := LBS(\hat\mu_i^\text{stable})$$ by enforcing boundary conditions.
Then we warp $I_i^\text{samp}$ with $\overrightarrow{\hat{V}_i^\text{stable}}$ for each $i$ to obtain stabilized $I_i^{\text{stable}}$, \begin{equation}
I_i^{\text{stable}}= I^\text{samp}_i\boxplus \overrightarrow{\hat{V}_i^\text{stable}},\, 1 \leq i \leq T_\text{samp},
\end{equation}

Then we obtain a stabilized image sequence $\{ I_i^{\text{stable}} \}_{i=1}^{T_\text{samp}}$ in which distortion is suppressed.

\begin{figure}[t]

\centering
\begin{subfigure}[t]{0.15\textwidth}
\includegraphics[width=\textwidth]{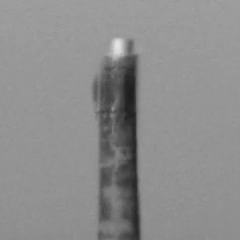}
\vspace{-1em}
\end{subfigure}
\begin{subfigure}[t]{0.15\textwidth}
\includegraphics[width=\textwidth]{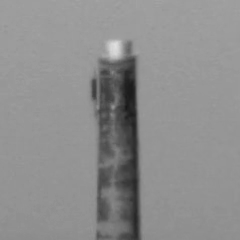}
\vspace{-1em}
\end{subfigure}
\begin{subfigure}[t]{0.15\textwidth}
\includegraphics[width=\textwidth]{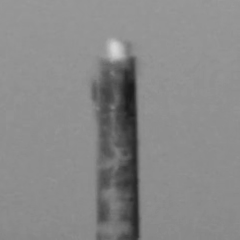}
\vspace{-1em}
\end{subfigure} \\
\begin{subfigure}[t]{0.15\textwidth}
\includegraphics[width=\textwidth]{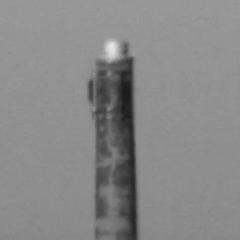}
\caption{}
\end{subfigure}
\begin{subfigure}[t]{0.15\textwidth}
\includegraphics[width=\textwidth]{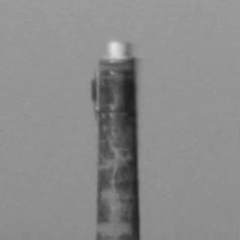}
\caption{}
\end{subfigure}
\begin{subfigure}[t]{0.15\textwidth}
\includegraphics[width=\textwidth]{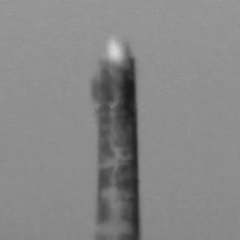}
\caption{}
\end{subfigure}
\caption{Stabilizing the sharp and large deformed frames in Chimney sequence. (a) Sharp and large deformed frame in original sequence. (b) Stabilized (a). (c) Replaced frame in stabilized sequence.}
\label{figure:chimney stable}
\end{figure}


\subsubsection{Absorbing stabilization}\label{subsubsec:AbsorbStabilization}
In the subsampling stage, we choose images which are both sharp and at most minorly distorted. As a result, some of the sharp images whose pixels are severely displaced are discarded. It would be useful if we can make use of those sharp but severely deformed video frames, because they may contain textures which are sharper than any Internal stabilized frame. However, we need to align them onto the same positions and shapes to fuse them. As those sharp frames are geometrically dissimilar to the reference image, estimating the deformation fields between them and the reference by optical flow would likely produce errors, especially if features are not detected in some frames due to occlusion or blurring. As such, warping them leads to large registration errors. On the other hand, we have acquired a stabilized video whose frames have little deformation from the reference image. Features are unlikely to be occluded or blurred out in all of these stabilized frames. Thus we can make use of the ample information in the stabilized sequence to stabilize the sharp and severely deformed images, and transform them to be minorly deformed. See Figure \ref{figure:chimney stable}.

\begin{algorithm}[t]
\begin{algorithmic}[1]
\REQUIRE Internal stabilized image sequence $\{ I_i^{\text{stable}} \}_{i=1}^{T_\text{samp}}$, original video $\{ I_i \}_{i=1}^{T}$
\ENSURE Absorbing stabilized image sequence $\{I_j^{\text{sharp}} \}_{j=1}^{T_\text{samp}}$
\STATE Obtain the set of the subsampled video frames $G$ and that of original sequence $H=\{h_j\}_{j=1}^{T_\text{samp}}$ with maximal sharpness
\STATE Obtain the intersection set $G \cap H$
\FOR{$j=1$ to $T_\text{samp}$}
\IF{$h_j \in H \setminus G $}
\STATE Stabilize $h_j$ with respect to the Internal stabilized sequence by (\ref{equation:absorbing stabilization}) to give $I_j^\text{sharp}$
\ENDIF
\IF{$h_j\in G\cap H$}
\STATE Take $I_j^\text{sharp}$ to be the Internal stabilized frame of $h_j$
\ENDIF
\ENDFOR
\RETURN $\{I_j^\text{sharp} \}_{j=1}^{T_\text{samp}}$
\end{algorithmic}
\caption{Absorbing stabilization}
\label{algorithm:Absorbing Stabilization}
\end{algorithm}

Suppose we have an Internal stabilized sequence and the sharpest $T_\text{samp}$ frames in the original sequence, denoted by $H$. Denote the subsampled sequence by $G$. We aim to make use of as many sharp frames in $H$ as possible. We propose to replace the Internal stabilized frames whose corresponding subsampled frames are not in $H$ with frames in $H\setminus G$. The latter frames are not aligned well with $I^\text{ref}$. Their geometric distortions from $I^\text{ref}$ have to be suppressed prior to registration. We name this procedure Absorbing stabilization, as it absorbs sharp frames into the subsample.

Absorbing stabilization is carried out in a similar manner with Internal stabilization. The only difference is that Internal stabilization acts on an element of the subsample with respect to which it is stabilized, whereas the ``absorbed" sharp frames in $H$ are not elements of either the subsampled sequence or the stabilized sequence. Denoting the deformation field from sharp frame $h_j\in H$ to $I^\text{stable}_i$ by $\overrightarrow{W_j^i}$,
\begin{align}
\label{equation:absorbing stabilization}
\mathcal W_j:&=\begin{pmatrix}|&|& &|\\
\text{vec}(\overrightarrow{W_j^1})&\text{vec}(\overrightarrow{W_j^2})&\cdots&\text{vec}(\overrightarrow{W_j^{T_\text{samp}}})\\
|&|& &|\end{pmatrix},\\
\mathcal W_{j,1}:&=\text{Re}(\mathcal W_j),\, \mathcal W_{j,2}:=\text{Im}(\mathcal W_j),\\
\mathcal W_{j,p}&=\tilde{\mathcal L}^*_{j,p}+\tilde{\mathcal S}^*_{j,p}, (\tilde{\mathcal L}_{j,p}^*,\tilde{\mathcal S}_{j,p}^*)=\argmin\limits_{\substack{L,S\\L+S=\mathcal W_{j,p}}}\Vert L\Vert_*+\lambda\Vert S\Vert_1,\\
\tilde{\mathcal L}_{j,p}^*&=\begin{pmatrix}|&|& &|\\
\text{vec}(\overrightarrow{\tilde{L}_{j,p}^1})&\text{vec}(\overrightarrow{\tilde{L}_{j,p}^2})&\cdots&\text{vec}(\overrightarrow{\tilde{L}_{j,p}^{T_\text{samp}}})\\|&|& &|\end{pmatrix},\\
\overrightarrow{W_j^\text{sharp}}&=\begin{cases}\dfrac{1}{T_\text{samp}} \sum\limits_{i=1}^{T_\text{samp}}\overrightarrow{\tilde{L}_{i,p}^j}&\text{if }\Vert\tilde{\mathcal S}_{i,p}^*\Vert_0<\dfrac{XYT_\text{samp}}{2}\\
\dfrac{1}{T_\text{samp}} \sum\limits_{j=1}^{T_\text{samp}}\overrightarrow{W_{i,p}^j}&\text{otherwise}\end{cases},\\
\hat\mu_j^\text{sharp}&=\begin{cases}\mu_j^\text{sharp}&\text{ if }|\mu_j^\text{sharp}|<1\\
\dfrac{\mu_j^\text{sharp}}{|\mu_j^\text{sharp}|+\varepsilon}&\text{ otherwise}\end{cases},\\
\overrightarrow{\hat{W}^\text{sharp}_j}&=LBS(\hat\mu_j^\text{sharp}),\, I^{\text{sharp}}_j=h_j\boxplus\overrightarrow{\hat{W}^\text{sharp}_j}.
\end{align}

\subsubsection{Registration}
\label{subsubsection:Registration}

From subsection \ref{subsection:Frame sampling}, we obtain a reference image $I^{\text{ref}}$, whose geometric structure is similar to the underlying truth. Suppose now we have Absorbing stabilized sequence $\{ I^\text{sharp}_j \}_{j=1}^{T_\text{samp}}$ and $I^{\text{ref}}$. Note that the geometric deformation of the Absorbing stabilized sequence is now suppressed and thus the error of the registration can be reduced. Applying Large Displacement Optical Flow, we have vector fields $\{\overrightarrow{V_j^\text{ref}}\}_{j=1}^{T_\text{samp}}$ from each Absorbing stabilized frame $I_j^\text{sharp}$ to $I^\text{ref}$. By Beltrami equation (\ref{equation:beltrami}), we can calculate $\mu_j^\text{ref}$, the Beltrami representation of $\overrightarrow{V_j^\text{ref}}$ for each $j$. Then we restrict $\Vert \hat{\mu}_j^\text{ref} \Vert_{\infty} < 1$ by thresholding it, i.e. \begin{equation} \hat{\mu}_j^\text{ref}=\begin{cases}\mu_j^\text{ref}&\text{if }|\mu_j^\text{ref}|<1\\\dfrac{\mu_j^\text{ref}}{|\mu_j^\text{ref}|+\varepsilon}&\text{otherwise}\end{cases}.\end{equation} We can obtain $$\overrightarrow{\hat{V}^\text{reg}_j} := LBS(\hat\mu_j^\text{ref})$$ by enforcing boundary conditions.
Then we warp $I^\text{sharp}_i$ with $\overrightarrow{\hat{V}^\text{reg}_j}$ for each $j$ to obtain registered $I_j^\text{reg}$, \begin{equation}
I_j^{\text{reg}}= I^\text{sharp}_j\boxplus \overrightarrow{\hat{V}^\text{reg}_j}, 1 \leq j \leq T_\text{samp},
\end{equation}
Then we obtain a registered image sequence $\{ I_i^{\text{reg}} \}_{i=1}^{T_\text{samp}}$.

\subsection{Image fusion}
\label{subsection:Image fusion}
\subsubsection{Detail extraction}
\begin{figure}
\centering
\includegraphics[width=0.4\textwidth]{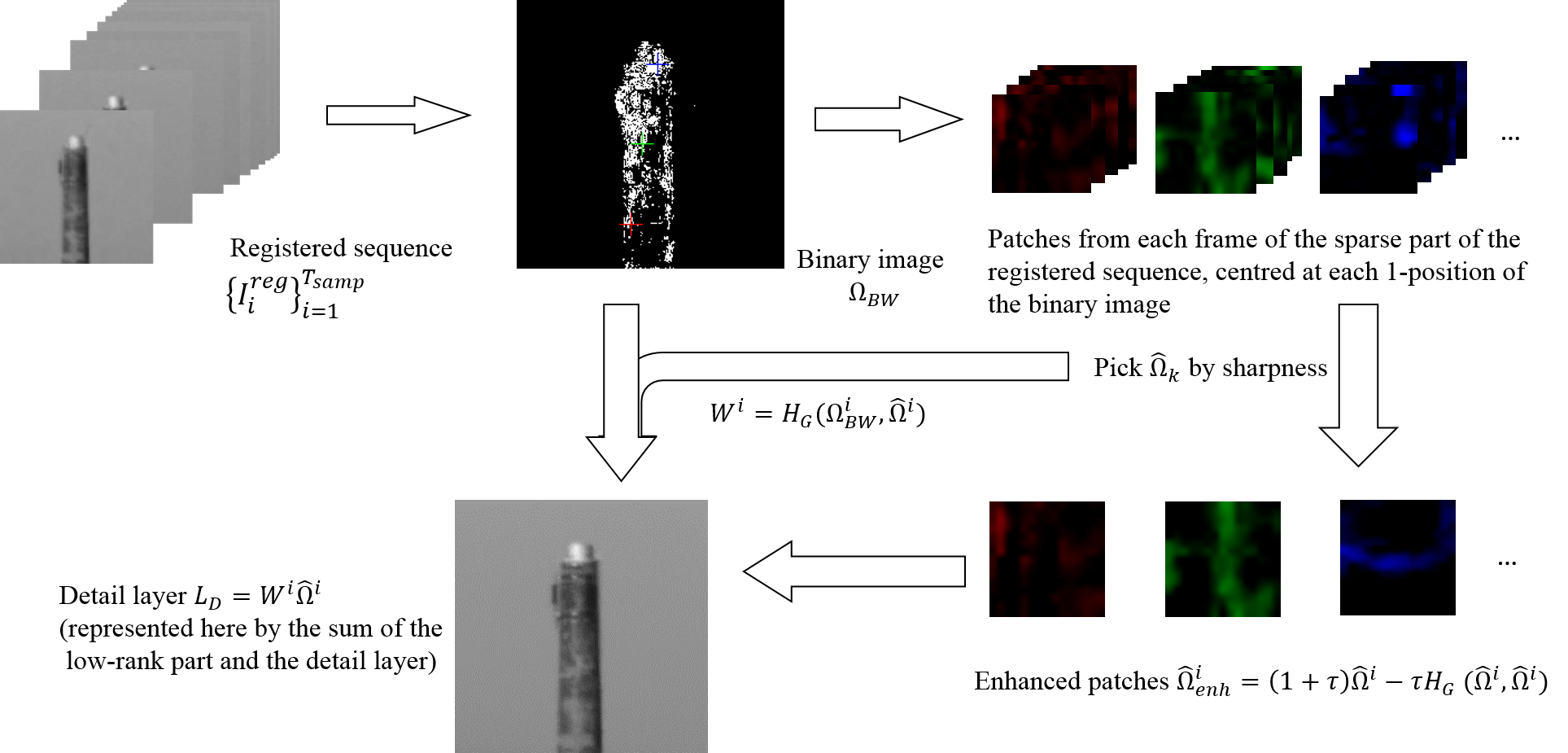}
\caption{Block diagram for Detail extraction.}
\label{figure:Fuse}
\end{figure}

From the above we now have a stabilized-then-registered sequence at hand. As there are blurry regions in each frame but at different spatial locations, we aim to extract information from each frame to produce an image as sharp and with as many texture details as possible. We adopt a modified version of the image fusion scheme from \cite{He2016}. Figure \ref{figure:Fuse} illustrates the process of Detail extraction.

\begin{figure}[t]
\begin{subfigure}[t]{0.15\textwidth}
\includegraphics[width=\textwidth]{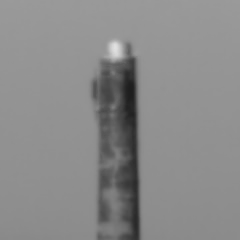}
\caption{}
\end{subfigure}\quad
\begin{subfigure}[t]{0.15\textwidth}
\includegraphics[width=\textwidth]{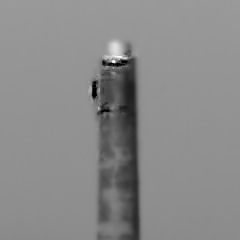}
\caption{}
\end{subfigure}\quad
\begin{subfigure}[t]{0.15\textwidth}
\includegraphics[width=\textwidth]{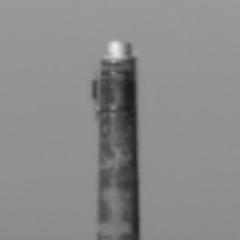}
\caption{}
\end{subfigure}
\caption{Detail extraction from \cite{He2016} and our modified version. (a) Low-rank part. (b) Summing sparse parts and adding onto low-rank part. (c) Taking maximum of absolute values (whenever overlapping) of sparse parts and adding onto low-rank part.}
\label{figure:maxfuse}
\end{figure}

\begin{figure}[t]
\centering
\begin{subfigure}[t]{0.15\textwidth}
\includegraphics[width=\textwidth]{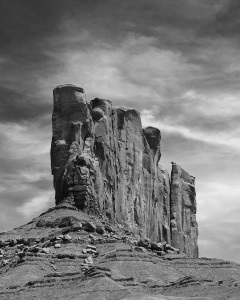}
\caption{}
\end{subfigure}
\begin{subfigure}[t]{0.15\textwidth}
\includegraphics[width=\textwidth]{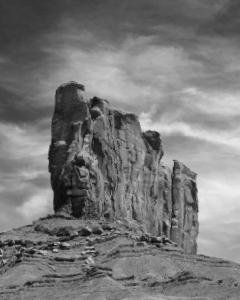}
\caption{}
\end{subfigure}
\begin{subfigure}[t]{0.15\textwidth}
\includegraphics[width=\textwidth]{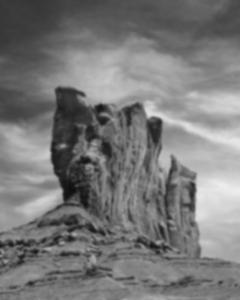}
\caption{}
\end{subfigure}
\\
\begin{subfigure}[t]{0.15\textwidth}
\includegraphics[width=\textwidth]{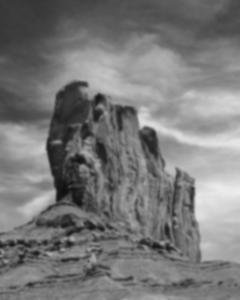}
\caption{}
\end{subfigure}
\begin{subfigure}[t]{0.15\textwidth}
\includegraphics[width=\textwidth]{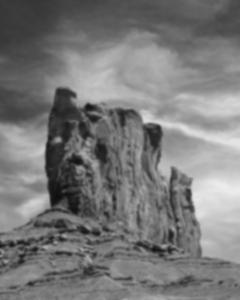}
\caption{}
\end{subfigure}
\begin{subfigure}[t]{0.15\textwidth}
\includegraphics[width=\textwidth]{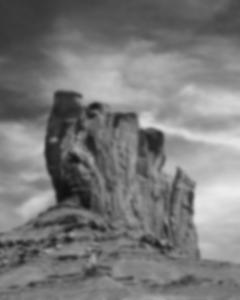}
\caption{}
\end{subfigure}
\caption{Original frames from Desert sequence. (a) Groundtruth. (b) A relatively clear and slightly distorted frame. (c,d,e,f) Distorted and blurry frames.}
\label{figure:simulated example}
\end{figure}

\begin{figure*}[t] 
\centering
\begin{subfigure}[t]{0.18\textwidth}
\includegraphics[width=\textwidth]{chimney_blur.png}
\vspace{-1em}
\end{subfigure}
\begin{subfigure}[t]{0.18\textwidth}
\includegraphics[width=\textwidth]{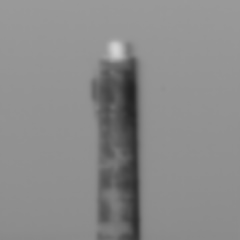}
\vspace{-1em}
\end{subfigure}
\begin{subfigure}[t]{0.18\textwidth}
\includegraphics[width=\textwidth]{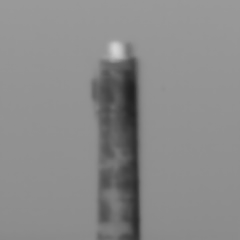}
\vspace{-1em}
\end{subfigure}
\begin{subfigure}[t]{0.18\textwidth}
\includegraphics[width=\textwidth]{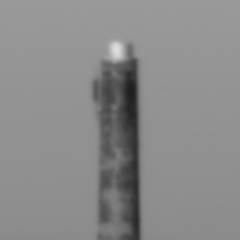}
\vspace{-1em}
\end{subfigure}
\begin{subfigure}[t]{0.18\textwidth}
\includegraphics[width=\textwidth]{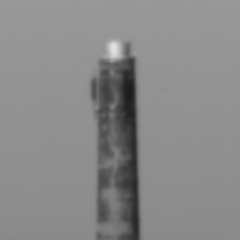}
\vspace{-1em}
\end{subfigure}
\\
\begin{subfigure}[t]{0.18\textwidth}
\includegraphics[width=\textwidth]{desert_original_frame_14.png}
\caption{}
\end{subfigure}
\begin{subfigure}[t]{0.18\textwidth}
\includegraphics[width=\textwidth]{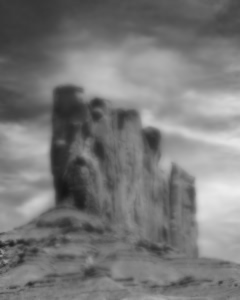}
\caption{}
\end{subfigure}
\begin{subfigure}[t]{0.18\textwidth}
\includegraphics[width=\textwidth]{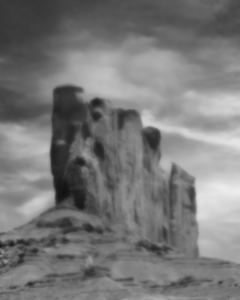}
\caption{}
\end{subfigure}
\begin{subfigure}[t]{0.18\textwidth}
\includegraphics[width=\textwidth]{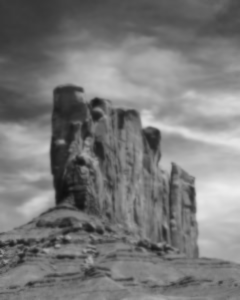}
\caption{}
\end{subfigure}
\begin{subfigure}[t]{0.18\textwidth}
\includegraphics[width=\textwidth]{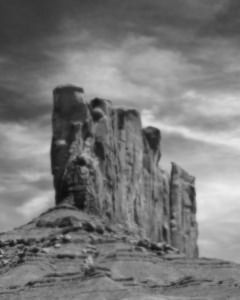}
\caption{}
\end{subfigure}
\caption{The reference images extracted from the Chimney and Desert sequences. (a) Observed. (b) Temporal mean. (c) Mean of low-rank by RPCA. (d) Centroid method \cite{Mario2014}. (e) Proposed method. The PSNR of (b), (c), (d), (e) from Desert sequence are 23.9529, 
24.6360, 26.0220, 27.4715 respectively. Note that blind deconvolution has not been applied to those results.}
\label{fig:reference image}
\end{figure*}

\begin{table*}[t!]
\centering
\caption{Performance of the restoration methods evaluated by PSNR and SSIM}
\label{tab:psnr ssim}

\begin{tabular}{|l || c | c | c | c | c| }
\hline
Sequence                 & SGL & Centroid & Two-stage & NDL & Proposed  \\ \hline
\hline
\multirow{2}{*}{Car}   & 21.1054 & 29.3143 & 26.1112 & 29.1757 & \textbf{31.7101}   \\ \cline{2-6}
                        & 0.7086 & 0.8842 & 0.7994 & 0.8703  & \textbf{0.9162}
\\ \hline
\hline
\multirow{2}{*}{Carfront}   & 16.7093 & 19.5172 & 15.3815 & 19.9009 & \textbf{24.0959}   \\ \cline{2-6}
                        & 0.6801 & 0.8163 & 0.5448 & 0.8136  & \textbf{0.9137}   \\ \hline
\hline
\multirow{2}{*}{Desert}   & 20.8255 & 27.7507 & 25.2696 & 22.9450 & \textbf{31.2749}   \\ \cline{2-6}
                        & 0.6985 & 0.8837 & 0.7886 & 0.7563  & \textbf{0.9385}   \\ \hline
\hline
\multirow{2}{*}{Road}   & 23.9782 & 30.0300 & 26.5800 & 27.4061 & \textbf{33.8682}   \\ \cline{2-6}
                        & 0.7638 & 0.8608 & 0.7827 & 0.8036  & \textbf{0.9063}   \\ \hline

\end{tabular}
\end{table*}

In \cite{He2016}, patches on the adaptive-thresholded sparse part are added together with weights to form a detail layer. The weights are obtained by smoothing the binary patches with guidance from their corresponding selected sharp patch, i.e.
\begin{equation}
W^i=H_G(\Omega_\text{BW}^i,\hat\Omega^i).
\end{equation}
\noindent where $H_G$ is a guided filter.

The detail layer is then obtained by the weighted sum
\begin{equation}
L^D=\sum\limits_{k=1}^MW^i\hat\Omega^i.
\end{equation}
Since the patch size $K$ can be large, the 1-positions in the binary image may be relatively close, and the patches may spatially overlap each other. As seen in our experiments, the fused detail intensities go out of bounds if there is overlap among the patches. See Figure \ref{figure:maxfuse}. As a remedy, instead of a weighted sum, we first multiply each patch with its corresponding weight $\tilde\Omega^i=W^i\hat\Omega^i$, and then pick for each spatial position the intensity with the highest absolute value amongst the patches, i.e.
\begin{equation}
L^D(x,y)=\tilde\Omega^{i_0}(x,y),
\end{equation}
where $i_0=\text{arg}\max\limits_i\{|\tilde\Omega^i(x,y)|\}$.
The detail layer $L^D$ with weight $\beta$ is held on to be fused in the next section.

\subsubsection{Deblurring}
\label{subsubsection:Deblurring}
While deblurring an image, the texture details are often overly sharpened, producing undesired artifacts. In accordance with \cite{He2016}, after RPCA is applied for detail extraction, the texture details are captured in the sparse part. Therefore, the low-rank part keeps the coarse structure and is blurry. Thus blind deconvolution \cite{Shan2008} is directly implemented on the low-rank part $I^{\text{LR}}$, which is then fused with the texture detail layer $L^D$
$$I^\text{final}=\text{deblur}(I^\text{LR})+\beta L^D.$$ Degradation caused by blur is generally modelled as follows,
\begin{equation}
G=F \otimes h+n,
\end{equation}
where $G$ is the blurred image, $F$ is the latent sharp image, h is the blur kernel. The blind deconvolution algorithm can be regarded as the following:
\begin{equation}\label{optimization deblur}
(\hat{F}, \hat{h}) = \textbf{argmin}_{F,h} \Vert Z-h \otimes F \Vert^2+\lambda_1 R_f(F)+\lambda_2 R_h(h),
\end{equation}
where $R_f$ and $R_h$ are the regularization terms used to restrain $F$ and $h$ based on their prior knowledge. 
The sparse regularization term in \cite{Shan2008} is defined as 
\begin{equation}
R_f(F)=\Vert\rho (F_x)+\rho (F_y)\Vert_1,
\end{equation}
where $F_x$ and $F_y$ are the image gradients of F in horizontal and vertical directions respectively, and $\rho(\cdot)$ is defined as \begin{equation}
\rho(x)=\begin{cases}
-\theta_1 |x| & x \leq l_t \\
-(\theta_2 x^2+\theta_3) & x>l_t.
\end{cases}
\end{equation} 
Here, $l_t,\theta_1, \theta_2, \theta_3$ are fixed parameters. Sparsity is also imposed to regularize the blur kernel $h$ as follows:
\begin{equation}
    R_h(h)=\Vert h\Vert_1.
\end{equation}
For the details for solving the optimization problem (\ref{optimization deblur}), we refer the reader to \cite{Shan2008}.

\section{Experimental Result and Discussion}
\label{section:Experimental Result and Discussion}

\begin{figure}[t]
\centering
\includegraphics[width=0.5\textwidth]{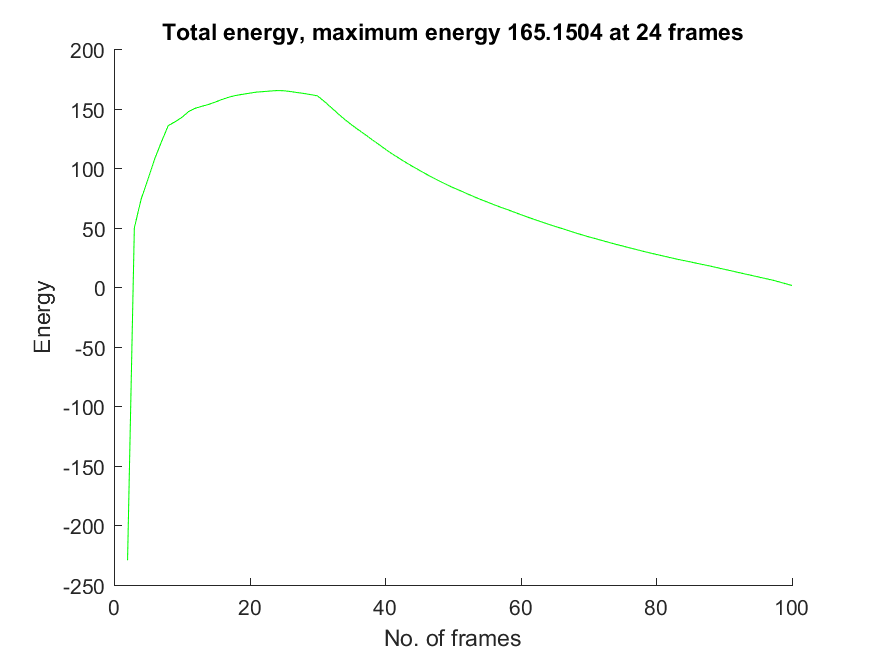}
\caption{The energy plot of the Desert sequence, with $\lambda = 200, \rho=0.1, \alpha=340$.}
\label{fig:energy_desert}
\end{figure}

\begin{figure}[t]
\centering
\begin{subfigure}[t]{0.24\textwidth}
\includegraphics[width=\textwidth]{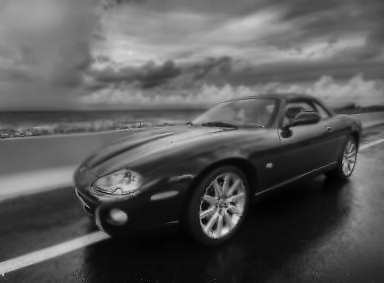}
\caption{}
\end{subfigure}
\begin{subfigure}[t]{0.24\textwidth}
\includegraphics[width=\textwidth]{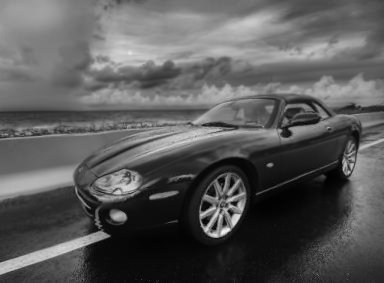}
\caption{}
\end{subfigure} \\
\begin{subfigure}[t]{0.15\textwidth}
\includegraphics[width=\textwidth]{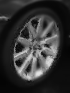}
\caption{}
\end{subfigure}
\begin{subfigure}[t]{0.15\textwidth}
\includegraphics[width=\textwidth]{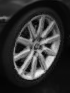}
\caption{}
\end{subfigure}
\caption{(a) is the fusion result of original Car sequence. (b) is the fusion result of the subsampled Car sequence. (c) and (d) are the zoomed part of (a) and (b) respectively. The PSNR of (a) and (b) are 28.4158 and 29.3634 respectively. Note that blind deconvolution has not been applied to those results.}
\label{fig:sub-sample importance}
\end{figure}

\begin{figure}[t]
\centering
\begin{subfigure}[t]{0.22\textwidth}
\includegraphics[width=\textwidth]{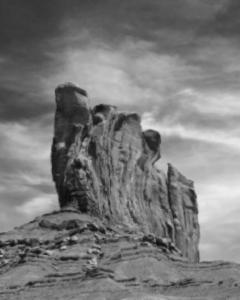}
\caption{}
\end{subfigure}
\begin{subfigure}[t]{0.22\textwidth}
\includegraphics[width=\textwidth]{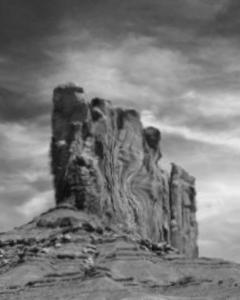}
\caption{}
\end{subfigure} \\
\begin{subfigure}[t]{0.22\textwidth}
\includegraphics[width=\textwidth]{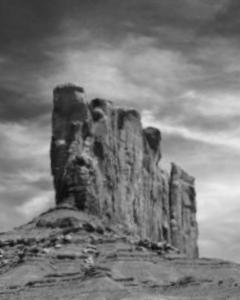}
\caption{}
\end{subfigure}
\begin{subfigure}[t]{0.22\textwidth}
\includegraphics[width=\textwidth]{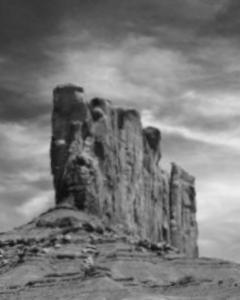}
\caption{} 
\end{subfigure} \\
\begin{subfigure}[t]{0.115\textwidth}
\includegraphics[width=\textwidth]{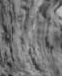}
\caption{zoomed (a)}
\end{subfigure}
\begin{subfigure}[t]{0.115\textwidth}
\includegraphics[width=\textwidth]{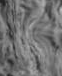}
\caption{zoomed (b)}
\end{subfigure} 
\begin{subfigure}[t]{0.115\textwidth}
\includegraphics[width=\textwidth]{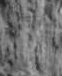}
\caption{zoomed (c)}
\end{subfigure}
\begin{subfigure}[t]{0.115\textwidth}
\includegraphics[width=\textwidth]{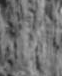}
\caption{zoomed (d)} 
\end{subfigure} \\
\begin{subfigure}[t]{0.115\textwidth}
\includegraphics[width=\textwidth]{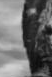}
\caption{zoomed (a)}
\end{subfigure}
\begin{subfigure}[t]{0.115\textwidth}
\includegraphics[width=\textwidth]{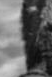}
\caption{zoomed (b)}
\end{subfigure} 
\begin{subfigure}[t]{0.115\textwidth}
\includegraphics[width=\textwidth]{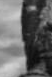}
\caption{zoomed (c)}
\end{subfigure}
\begin{subfigure}[t]{0.115\textwidth}
\includegraphics[width=\textwidth]{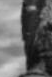}
\caption{zoomed (d)}
\end{subfigure}
\caption{(a) is the $17^\text{th}$ frame of original Desert sequence. (b) is the $17^\text{th}$ frame of symmetric constraint-based B-spline registered sequence \cite{Zhu2013}. (c) is the $17^\text{th}$ frame of stabilized Desert sequence. (d) is the $17^\text{th}$ frame of symmetric constraint-based B-spline registered sequence after stabilization.}
\label{fig:stabilization importance}
\end{figure}

\begin{figure*}[t!]
\centering
\begin{subfigure}[t]{0.24\textwidth}
\includegraphics[width=\textwidth]{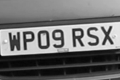}
\caption{Ground truth}
\end{subfigure}
\begin{subfigure}[t]{0.24\textwidth}
\includegraphics[width=\textwidth]{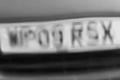}
\caption{A frame from the synthetic sequence}
\end{subfigure}
\begin{subfigure}[t]{0.24\textwidth}
\includegraphics[width=\textwidth]{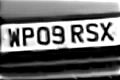}
\caption{Sobolev gradient-Laplacian method \cite{Lou2013}}
\end{subfigure}
\begin{subfigure}[t]{0.24\textwidth}
\includegraphics[width=\textwidth]{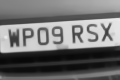}
\caption{Centroid method \cite{Mario2014}, deblurred with \cite{Shan2008}}
\end{subfigure}

\begin{subfigure}[t]{0.24\textwidth}
\includegraphics[width=\textwidth]{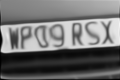}
\caption{Two-stage reconstruction method \cite{Oreifej2011}}
\end{subfigure}
\begin{subfigure}[t]{0.24\textwidth}
\includegraphics[width=\textwidth]{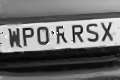}
\caption{Near-diffraction limited method \cite{Zhu2013}, deblurred with \cite{Shan2008}}
\end{subfigure}
\begin{subfigure}[t]{0.24\textwidth}
\includegraphics[width=\textwidth]{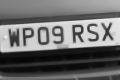}
\caption{Proposed method}
\end{subfigure}
\caption{Comparison of results on the Carfront sequence}
\label{carfront}
\end{figure*}

\begin{figure}[t]
\centering
\begin{subfigure}[t]{0.24\textwidth}
\includegraphics[width=\textwidth]{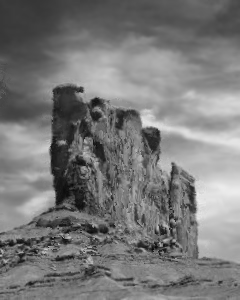}
\caption{}
\end{subfigure}
\begin{subfigure}[t]{0.24\textwidth}
\includegraphics[width=\textwidth]{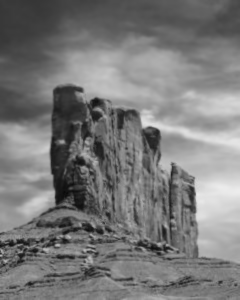}
\caption{}
\end{subfigure}
\caption{(a) is the NDL fusion result of symmetric constraint-based B-spline registered Desert sequence. (b) is the NDL fusion result of the stabilized symmetric constraint-based B-spline registered Desert sequence. The PSNR of (a) and (b) are 23.4055 and 27.8921 respectively. Note that blind deconvolution has not been applied to those results.}
\label{fig:stabilization importance fusion}
\end{figure}

In this section, detailed experimental justification of the proposed method will be illustrated. Firstly, we show the improvement of the reference image compared to several methods. Then, we show the importance of subsampling the video sequence, which not only obtains a better reference image but also reduces the computation time. Next, the advantages of stabilization are illustrated by comparing to those registering the frames alone. Finally, both qualitative and quantitative measures are used to evaluate the performance of the proposed algorithm comparing with several state-of-the-art methods. Peak Signal to Noise Ratio (PSNR) and Structural Similarity Index (SSIM) are computed to assess the quality of the restored images objectively.

For all experiments, the parameters $\lambda_\text{samp}$ and $\rho$ in the energy model (\ref{optimizationmodel}) in the subsampling stage are in the range of $[200, 500]$ and $0.1$ respectively. For the registration stage, the parameters of the Large Displacement Optical Flow \cite{optical flow 2009} used are the default setting. 
For our modified version of the image fusion scheme from \cite{He2016}, we perform adaptive thresholding by assigning 1 to entries whose absolute difference with the norm of the mean of the $7\times 7$ window centered at itself is greater than a threshold taken from the range $[0.5,2]$, and 0 elsewhere. The patch size $K$ is taken to be 7, and the weight $\tau$ in unsharp masking is taken to be 1.7. The fusion weight $\beta$ applied on the detailed layer $L^D$ is in the range of $[0,1]$. For the deblurring stage, we apply blind deconvolution to obtain the final output. Two separate sets of parameters are used for the synthetic image sequences and real sequences. In the synthetic experiments, the blur kernel is set to be $5 \times 5$, the noise level is chosen within the interval $[0.02,0.1]$, the deblurring weight is set to be 0.02 and all other parameters are set to default. The details of the parameters can be found in \cite{Shan2008} and its project page.\footnote{\url{http://www.cse.cuhk.edu.hk/leojia/programs/deblurring/deblurring.htm}} For real sequences, we set the same parameters to (7,9,0.03,0.2) to obtain the outputs. The proposed algorithm is implemented in Matlab with MEX and C++. All the experiments are executed on an Intel Core i7 3.4GHz computer.

\begin{figure}[t]
\centering
\begin{subfigure}[t]{0.24\textwidth}
\includegraphics[width=\textwidth]{car_fused_filtered.png}
\caption{}
\end{subfigure}
\begin{subfigure}[t]{0.24\textwidth}
\includegraphics[width=\textwidth]{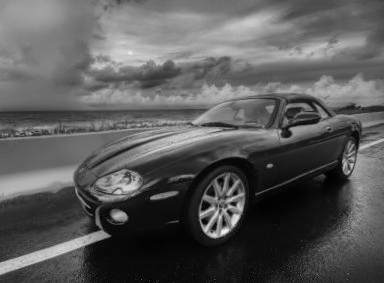}
\caption{}
\end{subfigure} \\
\begin{subfigure}[t]{0.08\textwidth}
\includegraphics[width=\textwidth]{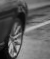}
\caption{}
\end{subfigure}
\begin{subfigure}[t]{0.08\textwidth}
\includegraphics[width=\textwidth]{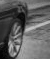}
\caption{}
\end{subfigure}
\begin{subfigure}[t]{0.15\textwidth}
\includegraphics[width=\textwidth]{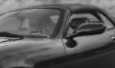}
\caption{}
\end{subfigure}
\begin{subfigure}[t]{0.15\textwidth}
\includegraphics[width=\textwidth]{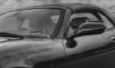}
\caption{}
\end{subfigure}
\caption{(a) is the fusion result of subsampled Car sequence. (b) is the fusion result of the stabilized Car sequence. (c), (e) and (d), (f) are the zoomed part of (a) and (b) respectively. The PSNR of (a) and (b) are 29.3634 and 30.3384 respectively. Note that blind deconvolution has not been applied to those results.}
\label{fig:stabilization importance fair test}
\end{figure}

\begin{figure*}[t]
\centering
\begin{subfigure}[t]{0.24\textwidth}
\includegraphics[width=\textwidth]{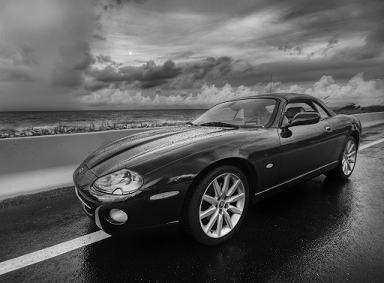}
\caption{Ground truth}
\end{subfigure}
\begin{subfigure}[t]{0.24\textwidth}
\includegraphics[width=\textwidth]{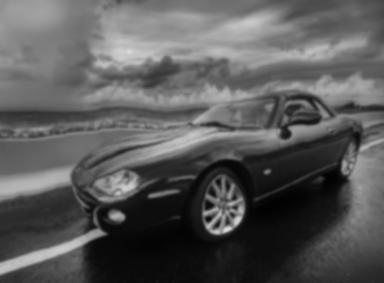}
\caption{A frame from the synthetic sequence}
\end{subfigure}
\begin{subfigure}[t]{0.24\textwidth}
\includegraphics[width=\textwidth]{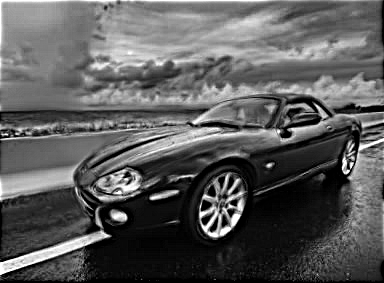}
\caption{Sobolev gradient-Laplacian method \cite{Lou2013}}
\end{subfigure}
\begin{subfigure}[t]{0.24\textwidth}
\includegraphics[width=\textwidth]{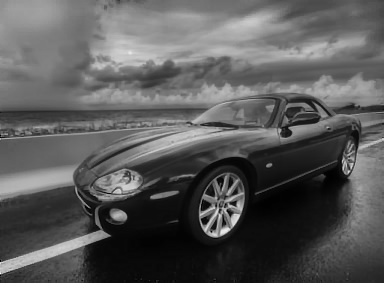}
\caption{Centroid method \cite{Mario2014}, deblurred with \cite{Shan2008}}
\end{subfigure}

\begin{subfigure}[t]{0.24\textwidth}
\includegraphics[width=\textwidth]{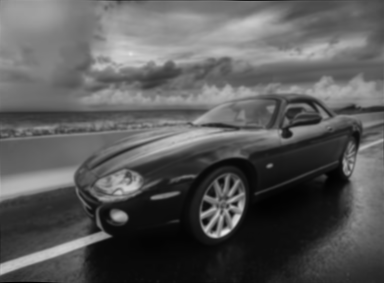}
\caption{Two-stage reconstruction method \cite{Oreifej2011}}
\end{subfigure}
\begin{subfigure}[t]{0.24\textwidth}
\includegraphics[width=\textwidth]{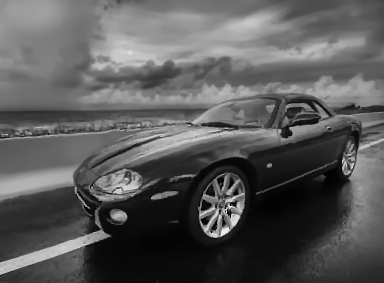}
\caption{Near-diffraction limited method \cite{Zhu2013}, deblurred with \cite{Shan2008}}
\end{subfigure}
\begin{subfigure}[t]{0.24\textwidth}
\includegraphics[width=\textwidth]{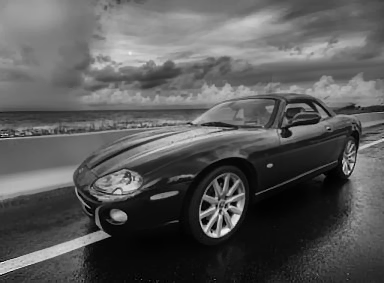}
\caption{Proposed method}
\end{subfigure}

\begin{subfigure}[t]{0.13\textwidth}
\includegraphics[width=\textwidth]{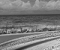}
\caption{Zoom-in of (a)}
\end{subfigure}
\begin{subfigure}[t]{0.13\textwidth}
\includegraphics[width=\textwidth]{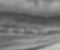}
\caption{Zoom-in of (b)}
\end{subfigure}
\begin{subfigure}[t]{0.13\textwidth}
\includegraphics[width=\textwidth]{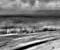}
\caption{Zoom-in of (c)}
\end{subfigure}
\begin{subfigure}[t]{0.13\textwidth}
\includegraphics[width=\textwidth]{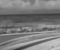}
\caption{Zoom-in of (d)}
\end{subfigure}
\begin{subfigure}[t]{0.13\textwidth}
\includegraphics[width=\textwidth]{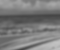}
\caption{Zoom-in of (e)}
\end{subfigure}
\begin{subfigure}[t]{0.13\textwidth}
\includegraphics[width=\textwidth]{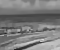}
\caption{Zoom-in of (f)}
\end{subfigure}
\begin{subfigure}[t]{0.13\textwidth}
\includegraphics[width=\textwidth]{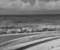}
\caption{Zoom-in of (g)}
\end{subfigure}

\begin{subfigure}[t]{0.13\textwidth}
\includegraphics[width=\textwidth]{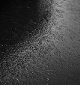}
\caption{Zoom-in of (a)}
\end{subfigure}
\begin{subfigure}[t]{0.13\textwidth}
\includegraphics[width=\textwidth]{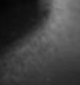}
\caption{Zoom-in of (b)}
\end{subfigure}
\begin{subfigure}[t]{0.13\textwidth}
\includegraphics[width=\textwidth]{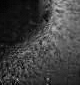}
\caption{Zoom-in of (c)}
\end{subfigure}
\begin{subfigure}[t]{0.13\textwidth}
\includegraphics[width=\textwidth]{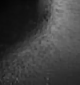}
\caption{Zoom-in of (d)}
\end{subfigure}
\begin{subfigure}[t]{0.13\textwidth}
\includegraphics[width=\textwidth]{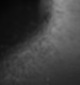}
\caption{Zoom-in of (e)}
\end{subfigure}
\begin{subfigure}[t]{0.13\textwidth}
\includegraphics[width=\textwidth]{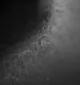}
\caption{Zoom-in of (f)}
\end{subfigure}
\begin{subfigure}[t]{0.13\textwidth}
\includegraphics[width=\textwidth]{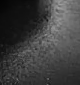}
\caption{Zoom-in of (g)}
\end{subfigure}
\caption{Comparison of results on the Car sequence}
\label{car}
\end{figure*}

The proposed method is compared with four representative methods: Sobolev gradient-Laplacian method \cite{Lou2013} (\textbf{SGL}), Centroid method \cite{Mario2014} (\textbf{Centroid}), the data-driven two-stage approach for
image restoration \cite{Oreifej2011} (\textbf{Two-stage}) and near-diffraction-limited-based image restoration for removing turbulence \cite{Zhu2013} (\textbf{NDL}).  For the Two-stage method, it was originally applied to restore a sharp image from an underwater video which was distorted by water waves. As videos degraded by water turbulence are generally treated as if under large distortion with mild blur, Two-stage still gets reasonable results and thus the comparison is valid. The codes of SGL \cite{Lou2013}, Two-stage \cite{Oreifej2011} and NDL \cite{Zhu2013} are provided by the respective authors, and the parameters used are default setting.

\begin{figure*}[t!]
\centering
\begin{subfigure}[t]{0.24\textwidth}
\includegraphics[width=\textwidth]{desert_truth.png}
\caption{Ground truth}
\end{subfigure}
\begin{subfigure}[t]{0.24\textwidth}
\includegraphics[width=\textwidth]{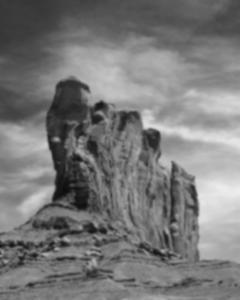}
\caption{A frame from the synthetic sequence}
\end{subfigure}
\begin{subfigure}[t]{0.24\textwidth}
\includegraphics[width=\textwidth]{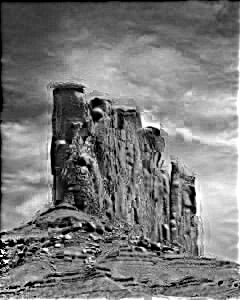}
\caption{Sobolev gradient-Laplacian method \cite{Lou2013}}
\end{subfigure}
\begin{subfigure}[t]{0.24\textwidth}
\includegraphics[width=\textwidth]{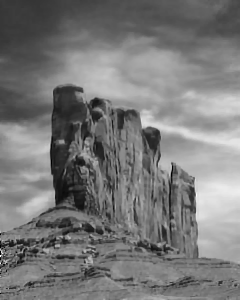}
\caption{Centroid method \cite{Mario2014}, deblurred with \cite{Shan2008}}
\end{subfigure}

\begin{subfigure}[t]{0.24\textwidth}
\includegraphics[width=\textwidth]{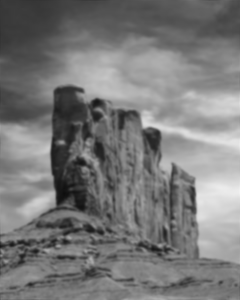}
\caption{Two-stage reconstruction method \cite{Oreifej2011}}
\end{subfigure}
\begin{subfigure}[t]{0.24\textwidth}
\includegraphics[width=\textwidth]{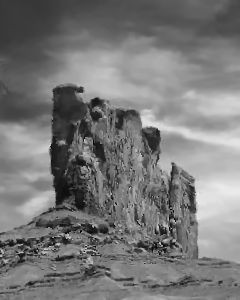}
\caption{Near-diffraction limited method \cite{Zhu2013}}
\end{subfigure}
\begin{subfigure}[t]{0.24\textwidth}
\includegraphics[width=\textwidth]{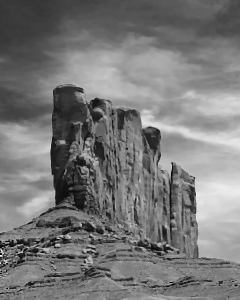}
\caption{Proposed method}
\end{subfigure}

\begin{subfigure}[t]{0.13\textwidth}
\includegraphics[width=\textwidth]{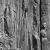}
\caption{Zoom-in of (a)}
\end{subfigure}
\begin{subfigure}[t]{0.13\textwidth}
\includegraphics[width=\textwidth]{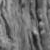}
\caption{Zoom-in of (b)}
\end{subfigure}
\begin{subfigure}[t]{0.13\textwidth}
\includegraphics[width=\textwidth]{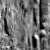}
\caption{Zoom-in of (c)}
\end{subfigure}
\begin{subfigure}[t]{0.13\textwidth}
\includegraphics[width=\textwidth]{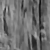}
\caption{Zoom-in of (d)}
\end{subfigure}
\begin{subfigure}[t]{0.13\textwidth}
\includegraphics[width=\textwidth]{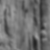}
\caption{Zoom-in of (e)}
\end{subfigure}
\begin{subfigure}[t]{0.13\textwidth}
\includegraphics[width=\textwidth]{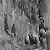}
\caption{Zoom-in of (f)}
\end{subfigure}
\begin{subfigure}[t]{0.13\textwidth}
\includegraphics[width=\textwidth]{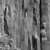}
\caption{Zoom-in of (g)}
\end{subfigure}

\begin{subfigure}[t]{0.13\textwidth}
\includegraphics[width=\textwidth]{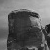}
\caption{Zoom-in of (a)}
\end{subfigure}
\begin{subfigure}[t]{0.13\textwidth}
\includegraphics[width=\textwidth]{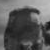}
\caption{Zoom-in of (b)}
\end{subfigure}
\begin{subfigure}[t]{0.13\textwidth}
\includegraphics[width=\textwidth]{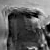}
\caption{Zoom-in of (c)}
\end{subfigure}
\begin{subfigure}[t]{0.13\textwidth}
\includegraphics[width=\textwidth]{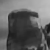}
\caption{Zoom-in of (d)}
\end{subfigure}
\begin{subfigure}[t]{0.13\textwidth}
\includegraphics[width=\textwidth]{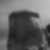}
\caption{Zoom-in of (e)}
\end{subfigure}
\begin{subfigure}[t]{0.13\textwidth}
\includegraphics[width=\textwidth]{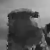}
\caption{Zoom-in of (f)}
\end{subfigure}
\begin{subfigure}[t]{0.13\textwidth}
\includegraphics[width=\textwidth]{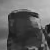}
\caption{Zoom-in of (g)}
\end{subfigure}
\caption{Comparison of results on the Desert sequence}
\label{desert}
\end{figure*}

\subsection{Quality of subsampled reference image} \label{subsec:Quality of sub-sampled reference image}

The visual quality of the reference images obtained by the proposed algorithm, temporal averaging, the temporal average of the low-rank and the centroid method \cite{Mario2014} are compared qualitatively in this subsection. The reference images are shown in Figure \ref{fig:reference image}: the first column (a) are the observed image from 'Chimney' and 'Desert' sequences while the other four columns are the reference images generated by temporal mean (b), mean of low rank (c), centroid method (d) and proposed algorithm (e). In the Chimney sequence, the subsampled reference image is sharper and preserves more details than the other three methods. This is because the subsampled sequence only consists of sharp and mildly distorted images, and hence the obtained reference image is clearer. For the other methods, the blurry and severely deformed frames are also taken into account so the reference image is corrupted. For an even more severely turbulence-degraded video (Desert sequence), the blurring effect is more noticeable. For the mean of the low-rank part, the general geometric structure is extracted and so sharp edges are preserved. However, most texture details will go to the sparse part, so the details are removed. For the centroid method, the texture details are kept as every image is warped by an average deformation field, and there is no direct manipulation on image intensities except interpolation. However, since the centroid method is based on the strong zero-mean assumption of the deformation fields between ground truth and the distorted sequence, which does not usually hold for turbulence-distorted video, the geometric structure may not be well kept. For the proposed method, the reference image is reconstructed from a good subsampled sequence, which minimizes the energy (\ref{optimizationmodel}) considering similarity and sharpness and improves iteratively. As a result, the edges are sharp, the geometric structure is preserved and the texture details are kept. The energy plot of (\ref{optimizationmodel}) is shown in Figure \ref{fig:energy_desert}. The PSNR of the reference images also justify the result.

\subsection{Importance of subsampling} \label{Importance of sub-sampling}

\begin{figure*}[t!]
\centering
\begin{subfigure}[t]{0.24\textwidth}
\includegraphics[width=\textwidth]{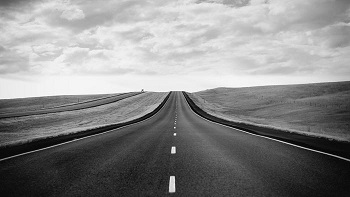}
\caption{Ground truth}
\end{subfigure}
\begin{subfigure}[t]{0.24\textwidth}
\includegraphics[width=\textwidth]{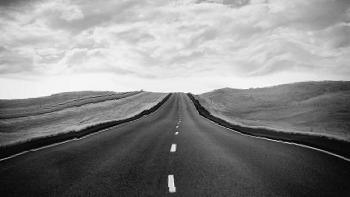}
\caption{A frame from the synthetic sequence}
\end{subfigure}
\begin{subfigure}[t]{0.24\textwidth}
\includegraphics[width=\textwidth]{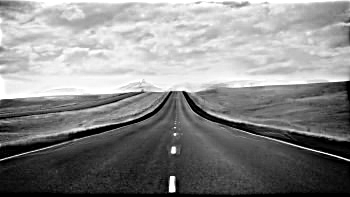}
\caption{Sobolev gradient-Laplacian method \cite{Lou2013}}
\end{subfigure}
\begin{subfigure}[t]{0.24\textwidth}
\includegraphics[width=\textwidth]{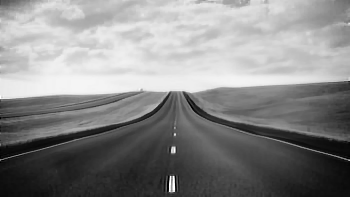}
\caption{Centroid method \cite{Mario2014}, deblurred with \cite{Shan2008}}
\end{subfigure}

\begin{subfigure}[t]{0.24\textwidth}
\includegraphics[width=\textwidth]{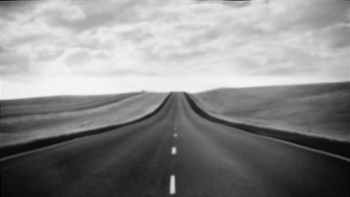}
\caption{Two-stage reconstruction method \cite{Oreifej2011}}
\end{subfigure}
\begin{subfigure}[t]{0.24\textwidth}
\includegraphics[width=\textwidth]{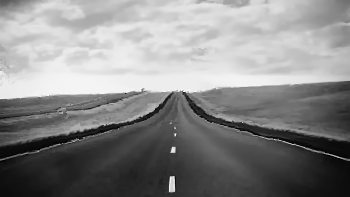}
\caption{Near-diffraction limited method \cite{Zhu2013}, deblurred with \cite{Shan2008}}
\end{subfigure}
\begin{subfigure}[t]{0.24\textwidth}
\includegraphics[width=\textwidth]{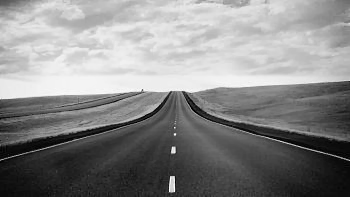}
\caption{Proposed method}
\end{subfigure}

\begin{subfigure}[t]{0.13\textwidth}
\includegraphics[width=\textwidth]{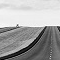}
\caption{Zoom-in of (a)}
\end{subfigure}
\begin{subfigure}[t]{0.13\textwidth}
\includegraphics[width=\textwidth]{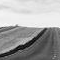}
\caption{Zoom-in of (b)}
\end{subfigure}
\begin{subfigure}[t]{0.13\textwidth}
\includegraphics[width=\textwidth]{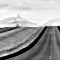}
\caption{Zoom-in of (c)}
\end{subfigure}
\begin{subfigure}[t]{0.13\textwidth}
\includegraphics[width=\textwidth]{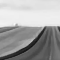}
\caption{Zoom-in of (d)}
\end{subfigure}
\begin{subfigure}[t]{0.13\textwidth}
\includegraphics[width=\textwidth]{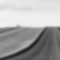}
\caption{Zoom-in of (e)}
\end{subfigure}
\begin{subfigure}[t]{0.13\textwidth}
\includegraphics[width=\textwidth]{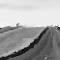}
\caption{Zoom-in of (f)}
\end{subfigure}
\begin{subfigure}[t]{0.13\textwidth}
\includegraphics[width=\textwidth]{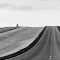}
\caption{Zoom-in of (g)}
\end{subfigure}

\begin{subfigure}[t]{0.13\textwidth}
\includegraphics[width=\textwidth]{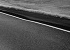}
\caption{Zoom-in of (a)}
\end{subfigure}
\begin{subfigure}[t]{0.13\textwidth}
\includegraphics[width=\textwidth]{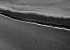}
\caption{Zoom-in of (b)}
\end{subfigure}
\begin{subfigure}[t]{0.13\textwidth}
\includegraphics[width=\textwidth]{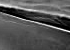}
\caption{Zoom-in of (c)}
\end{subfigure}
\begin{subfigure}[t]{0.13\textwidth}
\includegraphics[width=\textwidth]{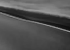}
\caption{Zoom-in of (d)}
\end{subfigure}
\begin{subfigure}[t]{0.13\textwidth}
\includegraphics[width=\textwidth]{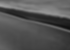}
\caption{Zoom-in of (e)}
\end{subfigure}
\begin{subfigure}[t]{0.13\textwidth}
\includegraphics[width=\textwidth]{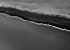}
\caption{Zoom-in of (f)}
\end{subfigure}
\begin{subfigure}[t]{0.13\textwidth}
\includegraphics[width=\textwidth]{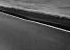}
\caption{Zoom-in of (g)}
\end{subfigure}
\caption{Comparison of results on the Road sequence}
\label{road}
\end{figure*}

In this subsection, we will illustrate the importance of subsampling. We compare the fusion results with and without subsampling. We register each video sequence to their corresponding reference image, which is the temporal mean of the sequence. Then fusion is applied to the two registered video sequences. Visual comparison and quantitative measures will be used to justify the result. Comparing (c) to (d) in Figure \ref{fig:sub-sample importance}, noticeable artifacts can be observed in the edges of the wheel and the overall image is also blurry. In contrast, comparing to the fusion result of the original video, the wheel in the subsampled sequence is free of artifacts, is sharper and has clearer edges. This observation can be explained by two factors: \begin{enumerate}
    \item Since the subsampled video is obtained by maximizing the energy that depends on the number of frames in the subsampled sequence, their similarity to the reference image, and their sharpness, the subsampled image frames mainly consist of comparatively sharp and less deformed image frames. Fewer noisy components are included in the sparse part in the fusion stage, and hence the result has a sharper edge and richer texture details are preserved.
    \item Since the reference image is constructed by a sharper and mildly distorted video sequence, the reference image is sharper and better preserves geometric structure. This has been justified in subsection \ref{subsec:Quality of sub-sampled reference image}. Therefore the alignments of the registered frames are more accurate, and the frames are thus more similar to the reference image. Therefore, the fusion artifacts due to poor registration become insignificant. 
\end{enumerate} 

\begin{figure*}[t!]
\centering
\begin{subfigure}[t]{0.24\textwidth}
\includegraphics[width=\textwidth]{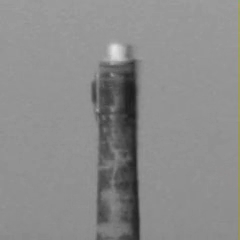}
\caption{An aligned and clear frame from the original sequence}
\end{subfigure}
\begin{subfigure}[t]{0.24\textwidth}
\includegraphics[width=\textwidth]{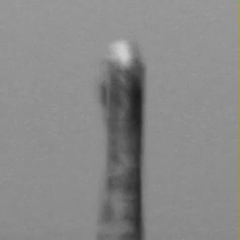}
\caption{A distorted and blurry frame from the original sequence}
\end{subfigure}
\begin{subfigure}[t]{0.24\textwidth}
\includegraphics[width=\textwidth]{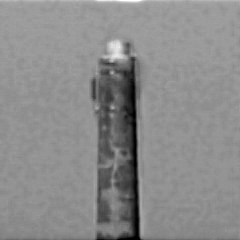}
\caption{Sobolev gradient-Laplacian method \cite{Lou2013}}
\end{subfigure}
\begin{subfigure}[t]{0.24\textwidth}
\includegraphics[width=\textwidth]{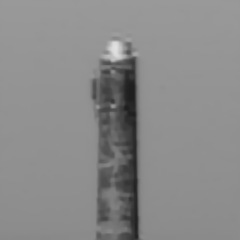}
\caption{Centroid method \cite{Mario2014}, deblurred with \cite{Shan2008}}
\end{subfigure}

\begin{subfigure}[t]{0.24\textwidth}
\includegraphics[width=\textwidth]{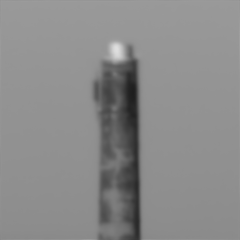}
\caption{Two-stage reconstruction method \cite{Oreifej2011}}
\end{subfigure}
\begin{subfigure}[t]{0.24\textwidth}
\includegraphics[width=\textwidth]{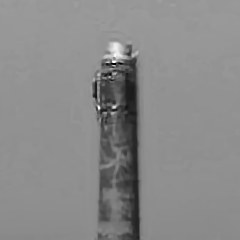}
\caption{Near-diffraction limited method \cite{Zhu2013}, deblurred with \cite{Shan2008}}
\end{subfigure}
\begin{subfigure}[t]{0.24\textwidth}
\includegraphics[width=\textwidth]{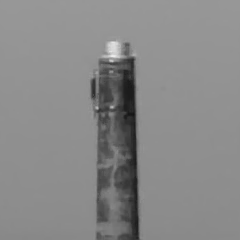}
\caption{Proposed method}
\end{subfigure}

\begin{subfigure}[t]{0.13\textwidth}
\includegraphics[width=\textwidth]{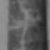}
\caption{Zoom-in of (a)}
\end{subfigure}
\begin{subfigure}[t]{0.13\textwidth}
\includegraphics[width=\textwidth]{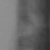}
\caption{Zoom-in of (b)}
\end{subfigure}
\begin{subfigure}[t]{0.13\textwidth}
\includegraphics[width=\textwidth]{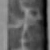}
\caption{Zoom-in of (c)}
\end{subfigure}
\begin{subfigure}[t]{0.13\textwidth}
\includegraphics[width=\textwidth]{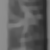}
\caption{Zoom-in of (d)}
\end{subfigure}
\begin{subfigure}[t]{0.13\textwidth}
\includegraphics[width=\textwidth]{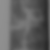}
\caption{Zoom-in of (e)}
\end{subfigure}
\begin{subfigure}[t]{0.13\textwidth}
\includegraphics[width=\textwidth]{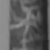}
\caption{Zoom-in of (f)}
\end{subfigure}
\begin{subfigure}[t]{0.13\textwidth}
\includegraphics[width=\textwidth]{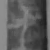}
\caption{Zoom-in of (g)}
\end{subfigure}

\begin{subfigure}[t]{0.13\textwidth}
\includegraphics[width=\textwidth]{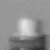}
\caption{Zoom-in of (a)}
\end{subfigure}
\begin{subfigure}[t]{0.13\textwidth}
\includegraphics[width=\textwidth]{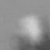}
\caption{Zoom-in of (b)}
\end{subfigure}
\begin{subfigure}[t]{0.13\textwidth}
\includegraphics[width=\textwidth]{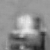}
\caption{Zoom-in of (c)}
\end{subfigure}
\begin{subfigure}[t]{0.13\textwidth}
\includegraphics[width=\textwidth]{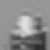}
\caption{Zoom-in of (d)}
\end{subfigure}
\begin{subfigure}[t]{0.13\textwidth}
\includegraphics[width=\textwidth]{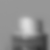}
\caption{Zoom-in of (e)}
\end{subfigure}
\begin{subfigure}[t]{0.13\textwidth}
\includegraphics[width=\textwidth]{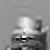}
\caption{Zoom-in of (f)}
\end{subfigure}
\begin{subfigure}[t]{0.13\textwidth}
\includegraphics[width=\textwidth]{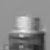}
\caption{Zoom-in of (g)}
\end{subfigure}

\caption{Comparison of results on the Chimney sequence}
\label{chimney}
\end{figure*}

\subsection{Importance of stabilization} \label{Advantages of stabilization}

This subsection demonstrates the importance of stabilization by applying low-rank decomposition on the deformation fields. Since stabilization is mainly used for enhancing the registration results, some fusion results will be shown, and the performance will be evaluated by visual comparison and PSNR. We will show the importance of the stabilization by comparing the fusion results with and without stabilization. Figure \ref{fig:stabilization importance fair test} shows the fusion result of the subsampled Car sequence with stabilization and that without stabilization, which is (b) in Figure \ref{fig:sub-sample importance}. The details are kept in a vivid way as the registration is more accurate and so the fusion is more satisfactory. Also, the details are sharper as absorbing stabilization is applied. (See zoomed parts in Figure \ref{fig:stabilization importance fair test}). The stabilization plays an important role in the proposed algorithm and makes a significant improvement because the geometric deformation of the video frames is further suppressed before registration. Therefore, the registration error can be reduced. Moreover, the Absorbing stabilization stabilizes the sharp but severely distorted frames. As a result, more texture details are kept in the fusion stage and thus the PSNR of the fusion result using stabilization is higher.   

On the other hand, the proposed stabilization scheme can be treated as a preprocessing step for registration. To illustrate this idea, we incorporate the stabilization scheme with the symmetric constraint-based B-spline registration proposed by Zhu and Milanfar \cite{Zhu2013}. This method is a common tool to tackle the turbulence-degraded video and gives satisfactory registration results in most cases. However, if the turbulence is strong in the sense that the frames are very blurry and severely distorted, the method may not obtain a good result.The registration results along with their corresponding fusion results are shown in Figure \ref{fig:stabilization importance} and Figure \ref{fig:stabilization importance fusion}. As shown in Figure \ref{fig:stabilization importance}, the registration results are improved significantly in the sense that the textures are not distorted (zoomed part (f) and (h)) and the edges are sharp (zoomed part (j) and (l)) after applying stabilization. As a result, the fusion result has also shown a significant improvement by stabilization in Figure \ref{fig:stabilization importance fusion}.

\begin{figure*}[t!]
\centering
\begin{subfigure}[t]{0.24\textwidth}
\includegraphics[width=\textwidth]{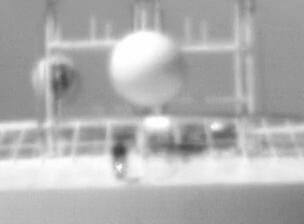}
\caption{A frame from the original sequence}
\end{subfigure}
\begin{subfigure}[t]{0.24\textwidth}
\includegraphics[width=\textwidth]{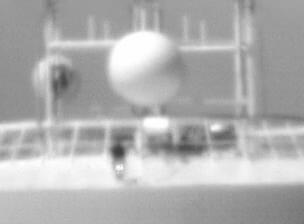}
\caption{Another frame from the original sequence}
\end{subfigure}
\begin{subfigure}[t]{0.24\textwidth}
\includegraphics[width=\textwidth]{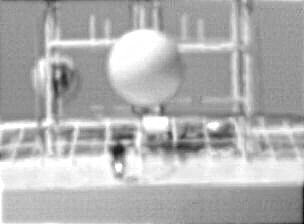}
\caption{Sobolev gradient-Laplacian method \cite{Lou2013}}
\end{subfigure}
\begin{subfigure}[t]{0.24\textwidth}
\includegraphics[width=\textwidth]{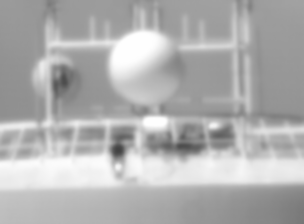}
\caption{Centroid method \cite{Mario2014}, deblurred with \cite{Shan2008}}
\end{subfigure}

\begin{subfigure}[t]{0.24\textwidth}
\includegraphics[width=\textwidth]{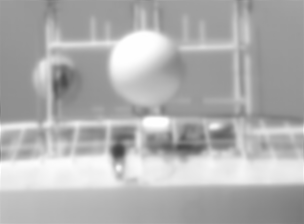}
\caption{Two-stage reconstruction method \cite{Oreifej2011}}
\end{subfigure}
\begin{subfigure}[t]{0.24\textwidth}
\includegraphics[width=\textwidth]{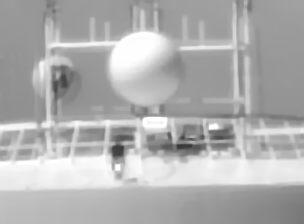}
\caption{Near-diffraction limited method \cite{Zhu2013}, deblurred with \cite{Shan2008}}
\end{subfigure}
\begin{subfigure}[t]{0.24\textwidth}
\includegraphics[width=\textwidth]{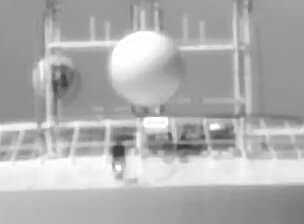}
\caption{Proposed method}
\end{subfigure}

\begin{subfigure}[t]{0.13\textwidth}
\includegraphics[width=\textwidth]{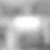}
\caption{Zoom-in of (a)}
\end{subfigure}
\begin{subfigure}[t]{0.13\textwidth}
\includegraphics[width=\textwidth]{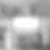}
\caption{Zoom-in of (b)}
\end{subfigure}
\begin{subfigure}[t]{0.13\textwidth}
\includegraphics[width=\textwidth]{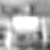}
\caption{Zoom-in of (c)}
\end{subfigure}
\begin{subfigure}[t]{0.13\textwidth}
\includegraphics[width=\textwidth]{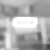}
\caption{Zoom-in of (d)}
\end{subfigure}
\begin{subfigure}[t]{0.13\textwidth}
\includegraphics[width=\textwidth]{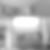}
\caption{Zoom-in of (e)}
\end{subfigure}
\begin{subfigure}[t]{0.13\textwidth}
\includegraphics[width=\textwidth]{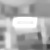}
\caption{Zoom-in of (f)}
\end{subfigure}
\begin{subfigure}[t]{0.13\textwidth}
\includegraphics[width=\textwidth]{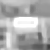}
\caption{Zoom-in of (g)}
\end{subfigure}

\begin{subfigure}[t]{0.13\textwidth}
\includegraphics[width=\textwidth]{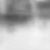}
\caption{Zoom-in of (a)}
\end{subfigure}
\begin{subfigure}[t]{0.13\textwidth}
\includegraphics[width=\textwidth]{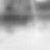}
\caption{Zoom-in of (b)}
\end{subfigure}
\begin{subfigure}[t]{0.13\textwidth}
\includegraphics[width=\textwidth]{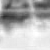}
\caption{Zoom-in of (c)}
\end{subfigure}
\begin{subfigure}[t]{0.13\textwidth}
\includegraphics[width=\textwidth]{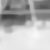}
\caption{Zoom-in of (d)}
\end{subfigure}
\begin{subfigure}[t]{0.13\textwidth}
\includegraphics[width=\textwidth]{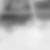}
\caption{Zoom-in of (e)}
\end{subfigure}
\begin{subfigure}[t]{0.13\textwidth}
\includegraphics[width=\textwidth]{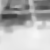}
\caption{Zoom-in of (f)}
\end{subfigure}
\begin{subfigure}[t]{0.13\textwidth}
\includegraphics[width=\textwidth]{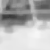}
\caption{Zoom-in of (g)}
\end{subfigure}

\caption{Comparison of results on the Water Tower sequence}
\label{watertower}
\end{figure*}

\subsection{Simulated Experiment} \label{Simulated Experiment}
To quantitatively evaluate the performance of the proposed algorithm, several sets of video sequences (namely Desert and Road) are generated with severe simulated turbulence distortions. Each frame of the simulated sequences is generated from a single image by randomly selecting $\frac{\text{width}\times \text{height}}{250} $ positions, and considering an image patch centered at each chosen position. A uniform motion vector patch with the same size of the image patch is generated, whereas the vector is randomly generated from a normal distribution for 2-vectors. Each vector patch is then smoothed with a Gaussian kernel and entrywise multiplied with a distorting strength value. The overall motion vector field is then generated by adding up the vector patches wherever overlapping. The image is then warped by the generated motion vector field. Note that the distortion effects are accumulated where the patches overlap. For each image frame, a Gaussian blur is applied to make them blurry. In the simulated experiments, the chosen patch size is $65 \times 65$, and the mean of the Gaussian kernel is slightly shifted for each image patch. See Figure \ref{figure:simulated example}. The Desert and Road sequences consist of $100$ frames each, among which 70 frames are degraded under severe distortion and the rest are deformed relatively mildly. The distorting strengths of severely distorted frames are in the range of $[1,1.5]$ while those of mildly distorted frames are in $[0.2,0.3]$. The Carfront sequence is a data set obtained from \cite{CLEAR2013} which contains mildly distorted frames when compared with the Desert and Road sequences. Note that the Carfront sequence is cropped from the original sequence. The Car sequence contains $80$ frames, among which only 15 are mildly distorted frames and the others are severely distorted. The distorting strengths of the mildly distorted frames and the severely distorted frames are in the ranges of $[0.3,0.5]$ and $[1,1.5]$ respectively. It serves as an extreme test case where most of the frames are severely degraded. \ref{tab:psnr ssim} gives the PSNR and SSIM values for all restoration results of five different restoration algorithms. Each sequence has two rows, where the first row denotes the PSNR values and the second row denotes the SSIM values.

\subsubsection{Mildly distorted sequences}
The Carfront sequence contains mildly distorted frames only and all the turbulence strength of images are similar. The restoration results of the Carfront sequence are shown in Figure \ref{carfront}. Since the deformations among Carfront frames are small, the restoration result of Centroid method and proposed algorithm are comparable. For the Carfront sequence, Centroid keeps the geometric structure well but the result is blurry. The shape of the restored image by SGL is slightly distorted and the intensities are unnatural. Two-stage distorts the image in a ripple-like pattern. NDL also keeps the structure relatively well but some artifacts are produced. The proposed algorithm keeps both the geometric structure and local details well.

\subsubsection{Strongly distorted sequences}
The majority of the frames in the Desert and Road sequences are strongly distorted, whereas the remaining are mildly distorted. The restoration results of the sequences are shown in Figure \ref{desert} and Figure \ref{road}. Since the deformations among Desert and Road frames are large, the restoration result of the proposed algorithm differs from existing methods. As the imaged objects are significantly displaced across frames, the temporal smoothing effect of the centroid method produces noticeable blur. This is more observable in the Desert experiment, where the many vertical edges are obscured by the blur, whereas in the Road sequence, thin strips parallel to the road are also diminished. A similar temporal smoothing effect manifests in SGL as overlapping shadowy artifacts. Intensity overshoots and jagged edges are observed in the results by NDL, likely because symmetric constraint-based B-spline registration cannot handle random discontinuous displacements across the temporal domain. In comparison, the proposed algorithm preserves clear edges and texture details. It is because the mildly distorted and sharp frames are selected and a good reference iamge is obtained in the subsampling stage. As a result, the proposed algorithm outperforms existing methods. 

\subsubsection{Extreme case: severely distorted sequence}
Most of the frames in the Car sequence are severely distorted, even more so than the Desert and Road sequences. Moreover, the distortions of the mildly distorted frames in the Car sequence are stronger than those in the Desert and Road sequences. The restoration results are shown in Figure \ref{car}. The result produced by the centroid method is fairly blurry, and its intensity contrast is significantly lower than other methods. Besides the intensity overshoots, several regions of the SGL result are noticeably deformed. The NDL result has fewer deformed regions, but it is relatively blurry, and straight edges are not preserved as well as the other methods. The proposed algorithm preserves geometric structure well and produces a clear image with minimal artifacts. The reason of the proposed algorithm outperforming the other methods in this extreme case is that the stabilization stage further suppresses the geometric deformation of the subsampled sequence. Therefore, a better registration result is achieved.  

\subsection{Real experiments}
\label{Real experiments}

We have also tested our proposed method on two real turbulence-distorted sequences, namely the Chimney and Water Tower sequences. A point of interest of real sequences is that among the two detriments of atmospheric distortion, blur is much more prominent than geometric deformation. The restoration results of the sequences are shown in Figures \ref{chimney} and \ref{watertower}.

As texture details in real sequences are mostly blurred, the ability of reconstruction schemes to extract sharp details is particularly crucial. In the presence of severely blurred frames, the Two-stage method cannot preserve texture details of both sequences and produces blurry results, as seen in Figure \ref{chimney}(l),(s) and Figure \ref{watertower}(l),(s). The temporal averaging in the centroid method also smooths out edges and sharp features as seen in Figure \ref{chimney}(k),(r). In this aspect, the Sobolev gradient-Laplacian method and near-diffraction limited method performs better, and reconstructs results with sharp details. However, due to varied reasons, the overall intensity distribution of their results differ from that of the original sequence. As a result, the pixel intensities of their results look unnatural. This is exemplified by the presence of dark strips in Figure \ref{chimney}(j),(m),(q),(t) and Figure \ref{watertower}(j),(q). The proposed algorithm produces images with relatively clear details, and preserves the intensity distribution of the original frames.

On the other hand, the severity of geometric deformation in real sequences cannot be underestimated, as exemplified by Figure \ref{chimney}(b). An ideal reconstruction scheme must accurately resolve such distortion. In lieu of an absent ground truth image, we compare the reconstructed results to the satisfactory original frame Figure \ref{chimney}(a). The near-diffraction limited method result show noticeable structure difference from Figure \ref{chimney}(a). The zoomed part Figure \ref{chimney}(m) shows vertical stretching, and Figure \ref{chimney}(t) highlights an additional kink on the right. The Two-stage reconstructed result is too blurry to identify the geometric structure within. In comparison, our proposed method preserves the geometric structure as well as the Sobolev gradient-Laplacian method and the centroid method.

In both of the above aspects, the proposed method is among the top performers.

\section{Conclusion}
\label{section:Conclusion}
The proposed algorithm produces better-aligned images compared to existing schemes when geometric deformation is severe. In addition, depending on the purpose of the reader, our algorithm can be partially implemented to suit their needs. For instance, to extract a stabilized video sequence from a distorted video sequence, the Deformation Removal algorithm can be applied. If computational time is essential and the need for sharpness is relaxed, the references extracted in subsection \ref{subsection:Reference image extraction and sub-sampling} would suffice. This enables near real-time stabilization of geometrically deformed video. Moreover, the stabilization scheme can be used as a prepossessing of registration and obtain a better registration result. 

However, due to the severity of distortions in the observed frames, a feature-matching optical flow scheme is required to obtain reliable deformation fields. In addition, numerous optical flow operations are required to obtain deformation fields. As a result, the algorithm is computationally expensive. We encourage interested readers to improve the algorithm and suggest suitable optical flow schemes.

\section*{Acknowledgments}

The $384\times 283$ image used for generating the Car sequence is a resized version of the Car image retrieved from the RetargetMe benchmark for image retargeting by Rubinstein, Gutierrez, Sorkine-Hornung and Shamir. The Carfront sequence is retrieved from the project webpage of \cite{CLEAR2013}. The $240\times 300$ image used for generating the Desert sequence is retrieved from WallpapersWide.com, whereas the $700\times 394$ image used for generating the Road sequence is by Dave and Les Jacobs for Blend Images and Getty Images. The $240\times 240\times 99$ Chimney sequence is produced by Hirsch and Harmeling from Max Planck Institute for Biological Cybernetics, whereas the $304\times 224\times 79$ Water Tower sequence is produced by Prof. Mikhail A. Vorontsov from the Intelligent Optics Lab of the University of Maryland. Both are recovered from \cite{Zhu2013}'s project webpage. The code for RPCA is from \cite{lrsmaster}. The code for optical flow is from \cite{optical flow 2009}. The authors would like to thank the above persons for allowing them to use the video, pictures and algorithms in their experiments. Lok Ming Lui is supported by HKRGC GRF (Project ID: 402413).




\bibliographystyle{IEEEtran}
\end{document}